%% file: neurips_2026.tex
\documentclass{article}

\usepackage[preprint]{neurips_2026}
\input{math_commands.tex}

\usepackage{xcolor}

\usepackage[utf8]{inputenc} 
\usepackage[T1]{fontenc}    
\usepackage{hyperref}       
\usepackage{url}            
\usepackage{booktabs}       
\usepackage{amsfonts}       
\usepackage{nicefrac}       
\usepackage{microtype}      
\usepackage{xcolor}         
\usepackage{amsmath} 
\usepackage{graphicx}
\usepackage{enumitem}
\usepackage{algorithm, algorithmic}
\usepackage{subcaption}

\title{Action-Conditioned Risk Gating for Safety-Critical Control under Partial Observability}

\author{
Yushen Liu \\
University of Virginia \\
\And
Yin-Jen Chen \\
Google \\
\And
Ziyi Chen \\
University of Maryland, College Park \\
\And
Tao Wang \\
Stanford University \\
\And
Heng Huang \\
University of Maryland, College Park \\
\And
Xugui Zhou \\
Louisiana State University \\
\And
Yanfu Zhang \\
College of William and Mary
}

\begin{document}

\maketitle

\begin{abstract}
Many safety-critical control problems are modeled as risk-sensitive partially observable Markov decision processes, where the controller must make decisions from incomplete observations while balancing task performance against safety risk. Although belief-space planning provides a principled solution, maintaining and planning over beliefs can be computationally costly and sensitive to model specification in practical domains. We propose a lightweight risk-gated reinforcement learning approximation for risk-sensitive control under partial observability. The method constructs a compact finite-history proxy state and learns an action-conditioned predictor of near-term safety violation. This predicted candidate-action risk is used in two complementary ways: as a risk penalty during value learning, and as a decision-time gate that interpolates between optimistic and conservative ensemble value estimates. As a result, low-risk actions are evaluated closer to reward-seeking estimates, while high-risk actions are evaluated more conservatively. We evaluate the approach in two safety-critical partially observable domains: automated glucose regulation and safety-constrained navigation. Across adult and adolescent glucose-control cohorts, the method improves overall glycemic tradeoffs and substantially reduces runtime relative to a belief-space planning baseline. On Safety-Gym navigation benchmarks, it achieves a more favorable reward-cost balance than unconstrained RL and several standard safe-RL baselines. These results suggest that action-conditioned near-term risk can provide an effective local signal for approximate risk-sensitive POMDP control when full belief-space planning is impractical.
\end{abstract}

\section{Introduction}

Many real-world control problems are both safety-critical and partially observable, arising in settings where agents must act under noisy sensing, imperfect control, and uncertain environment dynamics~\cite{lauri2022pomdp, carr2023shielding}. In such settings, the agent must choose actions from incomplete observations while avoiding rare but costly failures. Examples include automated glucose regulation, where hidden meal absorption and delayed insulin dynamics shape future risk, and safety-constrained navigation, where latent disturbances or limited observability can turn seemingly good actions into unsafe ones. In both cases, high performance alone is insufficient: the controller must act conservatively when near-term hazard is elevated, yet remain effective enough to achieve the task.

A principled framework for such problems is the partially observable Markov decision process (POMDP) \cite{pomdp}, in which the agent maintains a belief over latent states and plans in belief space. In practice, however, belief-state control can be difficult to deploy in safety-critical settings \cite{shani2013survey}. Accurate belief updates require transition and observation models that are often unavailable, and online belief-space planning can remain computationally expensive even with approximate models \cite{silver2010pomcp}. Moreover, while a full belief state is the principled sufficient statistic for optimal POMDP control, it may be more information than is needed by a particular approximate controller in some safety-critical tasks. For the settings considered here, we focus on a narrower decision question: whether a candidate action is likely to cause a near-term safety violation. We therefore investigate whether an action-conditioned estimate of near-term hazard, computed from recent history, can provide enough safety-relevant information for effective control without explicitly maintaining a full belief state~\cite{kaelbling1998planning,yu2006approximate,carr2023shielding}. This suggests that, for controllers whose safety decisions are dominated by short-horizon constraint risk, estimating action-conditioned near-term hazard may provide a practical alternative to maintaining a full latent-state belief.

\begin{figure}[ht]
\centering
\begin{minipage}[t]{0.5\textwidth}
\vspace{0pt}
\textbf{Positioning.}
Our method targets safety-critical control under partial observability, such as automated glucose regulation and safety-constrained navigation. Unlike classical POMDP methods, it avoids explicit belief updates and online belief-space planning by using a compact proxy state. Unlike standard safe RL, which often assumes full state access, it estimates action-conditioned near-term risk from partial observations and uses this risk to interpolate between optimistic and conservative ensemble Q-values. This provides a lightweight middle ground between belief-space POMDP planning and fully observable safe RL.
\end{minipage}
\hfill
\begin{minipage}[t]{0.48\textwidth}
\vspace{0pt}
\centering
\includegraphics[width=\linewidth]{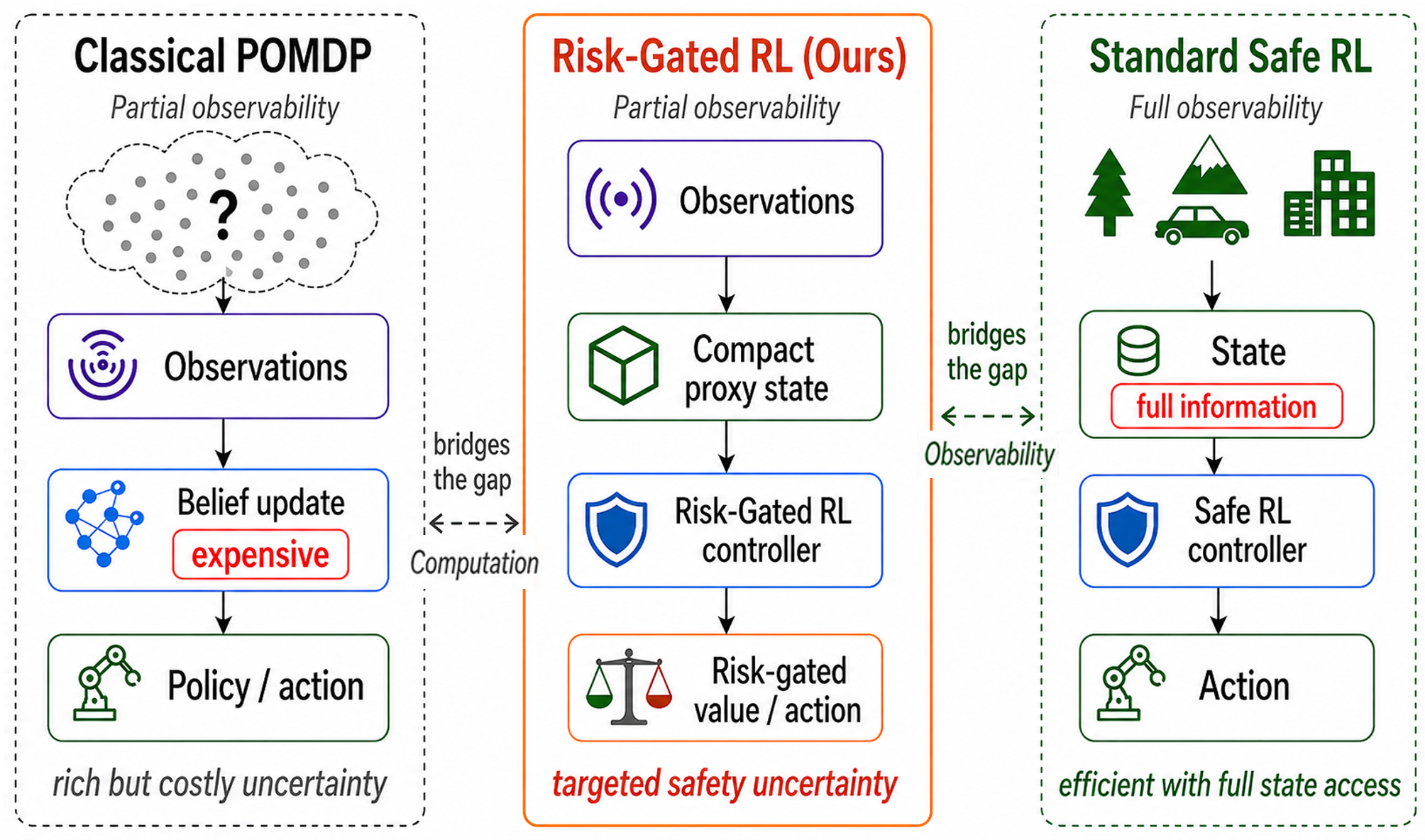}
\caption{Positioning of risk-gated RL relative to classical POMDPs and standard safe RL.}
\label{fig:positioning}
\end{minipage}
\end{figure}

This observation motivates a lighter alternative to explicit belief-space control. 
We propose a risk-gated reinforcement learning framework for safety-critical control under partial observability that avoids explicit belief updates by using a compact history-dependent proxy state and an action-conditioned estimate of near-term hazard. 
Figure~\ref{fig:positioning} summarizes how the proposed approach is positioned relative to classical POMDP planning and standard safe RL: it targets partial observability like POMDP methods, but avoids explicit belief updates, while retaining the computational simplicity of RL-style action evaluation.

The key design choice is to estimate near-term risk for each candidate action directly from recent history, rather than infer a full latent-state posterior\cite{roy2005belief,rigter2023risk}. For each candidate action \(a \in \mathcal{A}\), the controller predicts
\[
\hat\rho_t(a) \approx \mathbb{P}(E_t(a)=1 \mid H_t),
\]
where \(E_t(a)\) denotes a safety violation within a short horizon. This predicted hazard is then used to gate action evaluation between more optimistic and more conservative value estimates, so that the controller behaves more aggressively for low-risk actions and more conservatively for high-risk ones.

Our approach can be interpreted as a decision-oriented compressed-POMDP approximation. Instead of learning or updating a full latent belief, we retain two quantities aimed at action selection: a compressed summary of recent observations and an action-conditioned predictor of near-term hazard. This avoids explicit belief updates while aiming to retain the information most relevant to short-horizon safety-aware control. The resulting framework is lightweight, modular, and easy to instantiate in practical discretized action spaces. Additional intuition on the relation between the proposed method, latent-state POMDP views, and a surrogate risk-aware representation is provided in Appendix~\ref{app:pomdp_surrogate}.

We evaluate the proposed method in two different safety-critical partially observable settings: automated glucose regulation in the FDA-accepted UVa/Padova simulator\cite{FDA_simulator,zhou2022design}, and safe navigation in Safety-Gym benchmarks\cite{safety_gym}. These domains differ substantially in dynamics, observations, and task objectives, but share the same central challenge: the agent must act under partial observability while balancing performance and constraint avoidance. Across both domains, we find that risk-gated control yields a strong balance among performance, safety, and computational efficiency relative to both belief-based planning and standard safe-RL baselines.

This paper makes three main contributions. First, we introduce a framework for safety-critical partially observable control that avoids explicit belief updates by using a compact history-dependent proxy state and an action-conditioned estimate of near-term hazard. Second, we propose a risk-gated action-evaluation rule that uses predicted hazard to interpolate between optimistic and conservative value estimates. Third, we demonstrate across glucose regulation and Safety-Gym navigation that this design yields a favorable balance among task performance, safety, and computational efficiency.

\section{Related Work}

\paragraph{Partial observability and belief-based control.}
POMDPs provide a principled framework for sequential decision-making under partial observability by maintaining a belief state over latent variables\cite{pomdp}. In principle, belief-space planning can recover optimal decisions under uncertainty, but in practice exact filtering and planning are often intractable in continuous, high-dimensional, or safety-critical settings\cite{pineau2003pbvi,shani2013survey}. Approximate belief compression, learned latent-state models, and recurrent RL reduce this burden, but they still aim to preserve a broad summary of future-relevant uncertainty through either explicit belief maintenance or implicit hidden state\cite{kurniawati2008sarsop,igl2018deep}. 
These approaches aim to preserve information useful for general future prediction or long-horizon control.
Our work takes a more decision-oriented perspective. Rather than learning or updating a general-purpose belief or latent state, we focus on a narrower safety-relevant quantity: the action-conditioned probability of a near-term violation. This does not imply that full belief inference is unnecessary in general POMDPs. Instead, we study whether, in the safety-critical tasks considered here, a compact recent-history proxy together with an explicit near-term hazard estimate can provide a practical surrogate for decision-time safety assessment.

\paragraph{Safe reinforcement learning.}
Safe RL studies how to optimize performance while respecting constraints or limiting unsafe behavior during learning and deployment\cite{ray2019benchmarking}. Most safe-RL methods handle safety through expected costs, constraints, Lagrangian penalties, shielding, or model-based intervention\cite{achiam2017cpo,ibrahim2024reward,konighofer2025shields}, but they do not explicitly model how the near-term hazard of each candidate action varies under partial observability. Conversely, many partially observable RL methods learn hidden-state summaries without exposing an explicit action-conditioned safety quantity that can directly shape decision-time action evaluation. 
Our method is complementary to these approaches. It uses a learned near-term risk predictor not only as a penalty or constraint signal, but also as a gate that directly interpolates between optimistic and conservative value estimates for each candidate action. In this sense, predicted hazard affects how actions are evaluated at decision time, rather than only modifying the training objective or filtering actions after value estimation.

\input{wip}

\input{experiments}

\section{Conclusion}
We introduced a risk-gated reinforcement learning framework for safety-critical partially observable control. The method uses a compact history-dependent proxy state and an action-conditioned estimate of near-term hazard instead of explicit belief updates. Predicted risk shapes critic learning through a risk-penalized reward and gates decision-time action evaluation between optimistic and conservative ensemble values, allowing the controller to adapt its conservatism to each candidate action. Across glucose regulation and Safety-Gym navigation, the approach achieved favorable safety performance tradeoffs while reducing reliance on online belief-space planning. These results suggest that action-conditioned near-term risk can serve as a practical decision signal for approximate risk-sensitive control when full belief maintenance is computationally costly or difficult to specify.

\paragraph{Limitations.}
The main limitations are that the proxy representation and hazard features remain partially domain-specific, and the risk penalty and threshold parameters are manually calibrated. Future work should investigate unified learned representations and adaptive risk sensitivity.

\paragraph{Impact Statement.}
This work may have positive societal impact by supporting safer and more computationally efficient decision-making in safety-critical partially observable domains. Potential negative impacts include over-reliance on learned risk estimates or premature deployment without sufficient validation under distribution shift.



\bibliographystyle{plain}
\bibliography{references}


\appendix

\newpage
\section{POMDP Interpretation and Surrogate-View Intuition}
\label{app:pomdp_surrogate}

This appendix provides additional intuition for how the proposed method relates to partially observable control. The goal is not to replace the main formulation with a new theoretical framework, but to clarify how the method can be interpreted as a lightweight surrogate for belief-based decision-making. In particular, the appendix explains three ideas: (i) how the proposed controller relates to a latent-state POMDP view, (ii) how the risk-gated value rule can be interpreted as a surrogate long-horizon objective, and (iii) how approximation quality can be informally decomposed into representation error and risk-estimation error.

\subsection{From latent-state control to a risk-aware surrogate representation}
\label{app:latent_to_surrogate}

A partially observable Markov decision process (POMDP) is typically defined by the tuple
\[
(\mathcal{S}, \mathcal{A}, \mathcal{T}, \mathcal{R}, \mathcal{O}, \Omega, \beta),
\]
where \(s_t \in \mathcal{S}\) is the latent state, \(a_t \in \mathcal{A}\) is the action, \(o_t \in \mathcal{O}\) is the observation, \(\mathcal{T}\) is the transition model, \(\Omega\) is the observation model, and \(\beta \in (0,1)\) is the discount factor. In a classical POMDP solution, the agent maintains a belief state \(b_t\), that is, a posterior distribution over latent states conditioned on the full history.

In the safety-critical control settings considered in this paper, the full latent state may include hidden physiological variables, delayed actuation effects, unobserved disturbances, or other internal quantities that are not directly measurable. The classical solution is to update a belief over these latent quantities and plan in belief space. Our method instead adopts a lighter decision-oriented approximation.

Specifically, rather than maintaining a full belief state, the controller uses two quantities:
\[
o_t=\phi(\mathcal{W}_t^W),
\qquad
\hat\rho_t(a) \approx \mathbb{P}(E_t(a)=1 \mid H_t),
\]
where \(H_t\) is the interaction history, \(o_t\) is a compact proxy state derived from recent observations and actions, and \(\hat\rho_t(a)\) is a learned estimate of the probability that candidate action \(a\) will trigger a near-term safety violation within a fixed short horizon. Thus, the method does not attempt to reconstruct the full posterior over latent state. Instead, it retains only two components needed for action selection: a compressed summary of recent control context and an action-conditioned estimate of near-term hazard.

This viewpoint can be interpreted as a decision-oriented surrogate representation. The central idea is not that the pair \((o_t,\hat\rho_t(a))\) is sufficient for exact optimal control in a general POMDP. Rather, the claim is that in many safety-critical problems, much of the decision burden is concentrated in a smaller question: whether a candidate action appears likely to push the system toward a near-term unsafe regime. When this is true, a compact proxy state together with a near-term hazard estimate may retain the information most relevant for practical control, while avoiding the computational burden of explicit belief maintenance.

\subsection{Relation to belief compression}
\label{app:belief_compression}

The proposed framework is also related in spirit to approximate belief compression. In compressed-belief approaches, one seeks a lower-dimensional representation of the belief state that preserves enough information for effective planning. Our method can be viewed as an even more task-directed reduction: instead of compressing belief for general future prediction, it extracts information primarily for safety-aware action choice.

From this perspective, the predicted risk \(\hat\rho_t(a)\) acts as a decision-relevant statistic rather than a full latent summary. It does not attempt to represent all aspects of uncertainty in the environment. Instead, it targets one specific uncertainty quantity that is often critical in safety-aware control: whether executing action \(a\) is likely to cause a violation soon. This is why the framework is computationally lighter than classical belief-space control: it avoids recursive posterior updates and online belief-space planning, replacing them with direct prediction from recent history.

\subsection{Observable, latent, and predicted quantities}
\label{app:observable_latent_predicted}

It is useful to distinguish explicitly among three kinds of variables in the framework.

\paragraph{Latent state.}
At each time step \(t\), the environment has an underlying state \(s_t\) that is not directly observed. This state may include hidden physiological or environmental quantities that influence future reward and safety.

\paragraph{Observable history and proxy state.}
The agent receives observations and rewards and forms the history
\[
\mathcal{H}_t = \{(y_i,a_i,r_i)\}_{i\le t}.
\]
From this history it constructs a proxy state
\[
o_t=\phi(\mathcal{W}_t^W),
\]
which summarizes recent information relevant for short-horizon control and safety assessment.

\paragraph{Predicted near-term risk.}
For each candidate action \(a\), the controller predicts
\[
\hat\rho_t(a) \approx \mathbb{P}(E_t(a)=1 \mid H_t),
\]
where \(E_t(a)\) denotes the event that action \(a\) produces a safety violation within a short horizon. The controller then evaluates candidate actions using
\[
Q_t^{\mathrm{gate}}(a)
=
(1-\hat\rho_t(a))Q_t^+(a)
+
\hat\rho_t(a)Q_t^-(a),
\]
and selects actions using this mixed value, together with the risk threshold and fallback rules described in the main paper.

No explicit belief-state estimate \(b_t\) is computed. All learned components operate directly on recent history through \(o_t\) and \(\hat\rho_t(a)\).

\section{Interpretation of Risk-Gated Value Mixing}
\label{app:value_mixing}

The mixed value
\[
Q_t^{\mathrm{gate}}(a)
=
(1-\hat\rho_t(a))Q_t^+(a)
+
\hat\rho_t(a)Q_t^-(a)
\]
is the operational decision rule used by the controller. This appendix provides an interpretation of that rule.

Suppose \(Q_t^+(a)\) and \(Q_t^-(a)\) are optimistic and conservative evaluations of action \(a\), respectively. In the ensemble implementation used in the paper, these are obtained from the maximum and minimum values across an ensemble of critics. More generally, they can be understood as two action evaluations reflecting more reward-seeking and more safety-preserving viewpoints.

Then \(Q_t^{\mathrm{gate}}(a)\) can be interpreted as a risk-gated interpolation between these two modes of evaluation. When predicted risk is small, the controller places greater weight on the optimistic value and behaves more aggressively. When predicted risk is large, the controller shifts weight toward the conservative value and behaves more cautiously. Thus the risk signal does not merely act as a separate penalty term; it directly changes how future value is assessed at decision time.

This mixed value also admits a surrogate-objective interpretation. Informally, one may think of \(Q_t^+(a)\) and \(Q_t^-(a)\) as two long-horizon return estimates associated with more favorable and less favorable short-horizon safety outlooks. Under this view, the interpolation
\[
(1-\hat\rho_t(a))Q_t^+(a)+\hat\rho_t(a)Q_t^-(a)
\]
acts as a tractable proxy for a longer-horizon evaluation that conditions future desirability on predicted near-term hazard. We emphasize that this is an interpretation rather than a formal equivalence theorem: the paper does not claim that the mixed value exactly solves a surrogate MDP. Rather, the interpolation gives a practical mechanism for blending performance-seeking and safety-preserving action assessment in a manner that depends explicitly on the predicted risk of each candidate action.

\subsection{Connection to the shaped reward}
\label{app:shaped_reward_connection}

The main paper also introduces the shaped reward
\[
\tilde r_t = r_t - \lambda_{\mathrm{risk}}\hat\rho_t(a_t).
\]
This plays a complementary role to the mixed value. The mixed value affects action selection online by changing how candidate actions are ranked as a function of predicted hazard. The shaped reward affects learning by discouraging the critics from assigning high value to actions that repeatedly incur elevated predicted risk. Together, these two mechanisms encourage consistency between deployment-time conservatism and training-time value estimation.

\section{Theoretical Properties} \label{sec:theory}
 For simplicity, we consider a glucose regulation system as a motivating example to mathematically formulate a safe-gated POMDP.

\subsection{Notation and Definitions}

\subsubsection{Spaces and States}

\paragraph{Hidden State Space.} $s_t \in \mathcal{S} \subseteq \mathbb{R}^d$ represents the \emph{hidden physiological state} at time $t$, which is unobservable to the decision-making agent. This high-dimensional state encapsulates the complete physiological information relevant to glucose dynamics.

\paragraph{Observation function.} $o_t = \omega(s_t) \in \mathcal{G}$ denotes the \emph{observed glucose level}, where $\omega: \mathcal{S} \to \mathcal{G}$ is the physiological measurement mapping. We assume $\mathcal{G} \subseteq [g_{\min}, g_{\max}]$ is a bounded closed interval representing the feasible glucose measurement range. Throughout our main theoretical analysis, we assume this mapping is \emph{deterministic} (noiseless).

\paragraph{Action space.} $a_t \in \mathcal{A} \subset \mathbb{R}$ represents the \emph{action} (insulin dosage) administered at time $t$, where $\mathcal{A}$ is a bounded subset of the real line.

\paragraph{Reward function.} $r: \mathcal{S} \times \mathcal{A} \to \mathbb{R}$ is the \emph{reward function} that maps a physiological state-action pair to a scalar clinical reward. This function typically penalizes deviations of glucose levels from a target therapeutic range.

\paragraph{History trajectory.} $H_t = (o_0, a_0, o_1, a_1, \ldots, o_t) \in \mathcal{H}_t$ denotes the \emph{complete history trajectory} up to time $t$, where $\mathcal{H}_t$ represents the space of all possible histories of length $2t+1$.

\paragraph{Local observation window.} $\mathcal{W}_t^W = (o_{t-W+1}, a_{t-W+1}, \ldots, o_t) \in \mathcal{W}$ represents the \emph{local observation window} of fixed length $W \in \mathbb{N}^+$, constituting a finite suffix of $H_t$. This windowing mechanism provides a computationally tractable approximation to the full history.

\paragraph{Proxy state.} $\tilde{s}_t = \left(\mathcal{W}_t^W, \hat{\rho}_n(\mathcal{W}_t^W, \cdot)\right) \in \widetilde{\mathcal{S}}$ defines the \emph{proxy state}, which pairs the observation window with the empirical risk function. Here, $\hat{\rho}_n$ serves as shorthand for the mapping $a \mapsto \hat{\rho}_n(\mathcal{W}_t^W, a)$ when the window context is unambiguous.

\paragraph{Safety constraint.} $(g_{\mathrm{low}}, \infty) \subset \mathcal{G}$ defines the \emph{safe observation set} (target glucose range). The primary objective of our risk-gated policy framework is to ensure that the observed glucose level $o_t$ remains within the safe region $(g_{\mathrm{low}}, \infty)$ with high probability through appropriate action selection $a_t$.

\paragraph{Risk estimation framework.} We define the \emph{true tail risk} as the probability of hypoglycemia at the next step given the full history and proposed action: $\rho(H_t, a) := \mathbb{P}(\omega(s_{t+1}) < g_{\mathrm{low}} \mid H_t, a)$. Given a fixed clinical tolerance $\tau \in (0,1)$ as the \emph{safety threshold}, an action $a$ is considered safe at history $H_t$ if $\rho(H_t, a) \leq \tau$. To estimate this risk in practice, we employ an \emph{empirical risk estimator} trained on $n$ i.i.d. labelled samples $\{(\mathcal{W}^{(i)}, a^{(i)}, y^{(i)})\}_{i=1}^n$, where $y^{(i)} = \mathbf{1}_{\{\omega(s_{t+1}^{(i)}) < g_{\mathrm{low}} \}}$:

$$\hat{\rho}_n(\mathcal{W}_t^W, a) \;\approx\; \mathbb{E}\!\Bigl[\mathbf{1}_{\{\omega(s_{t+1}) < g_{\mathrm{low}} \}} \;\Big|\; \mathcal{W}_t^W,\, a\Bigr]$$

Since $\hat{\rho}_n$ estimates a probability, $\hat{\rho}_n \in [0,1]$ almost surely, and therefore $\sup \rho \leq 1$.

\paragraph{Policy.}

A \textbf{history-dependent policy} $\pi$ is a measurable map

$$\pi(\cdot \mid H_t):\; \mathcal{H}_t \;\to\; \Delta(\mathcal{A})$$

that assigns to each history a distribution over actions, where $\Delta(\mathcal{A})$ is the set of probability measures on $\mathcal{A}$. We write $a \sim \pi(\cdot \mid H_t)$ for an action drawn from this distribution, and denote the set of all admissible policies by $\Pi$.

The \textbf{policy risk} under $\pi$ at history $H_t$ is the expected per-step hypoglycemia probability:

$$\rho^\pi(H_t) \;:=\; \mathbb{E}_{a \sim \pi(\cdot \mid H_t)}\!\bigl[\rho(H_t, a)\bigr]$$

A policy $\pi$ is called safe if $\rho^\pi(H_t) \leq \tau$ for all $H_t$.

\textbf{Note on notation.} Throughout this appendix, we use $o_t = \omega(s_t) \in 
\mathcal{G}$ to denote the observed glucose level, which corresponds to $y_t$ in the 
main text. 

\subsubsection{Oracle Value Functions}
\paragraph{Value functions.} For a policy $\pi \in \Pi$, the state value function  $V^\pi$ measures the expected discounted cumulative reward starting from history $H_t$ and following $\pi$ thereafter:
$$V^\pi(H_t) \;:=\; \mathbb{E}_\pi\!\left[\sum_{k=0}^{\infty} \beta^k\, r(s_{t+k}, a_{t+k}) \;\Bigg|\; H_t\right]$$

where $\beta \in (0,1)$ is the discount factor and the expectation is taken over trajectories induced by $\pi$ and the environment dynamics.

\paragraph{Action-value function} $Q^\pi$. The expected discounted return when taking action $a$ at history $H_t$ and following $\pi$ thereafter:

$$Q^\pi(H_t, a) \;:=\; \mathbb{E}\bigl[r(s_t, a) \mid H_t, a\bigr] \;+\; \beta\, \mathbb{E}_{s_{t+1}}\!\bigl[V^\pi(H_{t+1}) \;\big|\; H_t, a\bigr]$$

The two functions are linked by $V^\pi(H_t) = \mathbb{E}_{a \sim \pi(\cdot \mid H_t)}[Q^\pi(H_t, a)]$.

\paragraph{Oracle safe action set} At each history $H_t$, the set of actions that satisfy the hard safety constraint under the true risk:

$$\mathcal{A}_{\mathrm{safe}}^*(H_t) \;=\; \bigl\{a \in \mathcal{A} \;\big|\; \rho(H_t, a) \leq \tau \bigr\}$$

\paragraph{Oracle Bellman operator} $\mathcal{T}: \mathcal{Q}(\mathcal{H} \times \mathcal{A}) \to \mathcal{Q}(\mathcal{H} \times \mathcal{A})$. The risk-gated Bellman operator restricts the greedy step to the safe action set:

$$(\mathcal{T} Q)(H_t, a) \;:=\; \mathbb{E}\bigl[r(s_t, a) \mid H_t, a\bigr] \;+\; \beta\, \mathbb{E}\!\left[\max_{a' \in \mathcal{A}_{\mathrm{safe}}^*(H_{t+1})} Q(H_{t+1}, a') \;\Bigg|\; H_t, a\right]$$

\paragraph{Oracle Q-function} $Q^*$ is the unique fixed point of $\mathcal{T}$ in $\bigl(\mathcal{Q}(\mathcal{H} \times \mathcal{A}),\, \|\cdot\|_\infty\bigr)$. The corresponding Oracle state value function and Oracle optimal policy are:

$$V^*(H_t) \;:=\; \max_{a \in \mathcal{A}_{\mathrm{safe}}^*(H_t)} Q^*(H_t, a), \qquad \pi^*(a \mid H_t) \;:=\; \operatorname*{arg\,max}_{a \in \mathcal{A}_{\mathrm{safe}}^*(H_t)} Q^*(H_t, a)$$

\subsubsection{Proxy Value Functions}
Since $H_t$ is unavailable at deployment time, we replace all Oracle objects with tractable proxy defined on the observed proxy state space $\widetilde{\mathcal{S}}$.

\paragraph{Proxy safe action set:}
$$\mathcal{A}_{\mathrm{safe}}(\tilde{s}_t) \;=\; \bigl\{a \in \mathcal{A} \;\big|\; \hat{\rho}_n(\mathcal{W}_t^W, a) \leq \tau \bigr\}$$

\paragraph{Proxy Bellman operator} $\widetilde{\mathcal{T}}: \mathcal{Q}(\widetilde{\mathcal{S}} \times \mathcal{A}) \to \mathcal{Q}(\widetilde{\mathcal{S}} \times \mathcal{A})$. The Lagrangian relaxation of the safety constraint, penalising constraint violations via the multiplier $\lambda$:

$$(\widetilde{\mathcal{T}}\,\widetilde{Q})(\tilde{s}_t, a) \;:=\; \mathbb{E}_{s \sim p(s \mid \mathcal{W}_t^W)}\!\bigl[r(s, a)\bigr] \;-\; \lambda\!\bigl(\hat{\rho}_n(\mathcal{W}_t^W, a) - \tau\bigr) \;+\; \beta\, \mathbb{E}\!\left[\max_{a' \in \mathcal{A}_{\mathrm{safe}}(\tilde{s}_{t+1})} \widetilde{Q}(\tilde{s}_{t+1}, a') \;\Bigg|\; \tilde{s}_t, a\right]$$

\paragraph{Proxy Q-function} $\widetilde{Q}^*$ is the unique fixed point of $\widetilde{\mathcal{T}}$. The corresponding proxy state value function and proxy policy are:

$$\widetilde{V}^*(\tilde{s}_t) \;:=\; \max_{a \in \mathcal{A}_{\mathrm{safe}}(\tilde{s}_t)} \widetilde{Q}^*(\tilde{s}_t, a), \qquad \tilde{\pi}^*(a \mid \tilde{s}_t) \;:=\; \operatorname*{arg\,max}_{a \in \mathcal{A}_{\mathrm{safe}}(\tilde{s}_t)} \widetilde{Q}^*(\tilde{s}_t, a)$$

\begin{remark}(Markov property of $\tilde{s}_t$) For $\widetilde{\mathcal{T}}$ to be a well-defined Bellman operator, $\tilde{s}_t$ must be a sufficient statistic for the one-step transition under the proxy model. This holds because $\hat{\rho}_n$ is a deterministic function of $(\mathcal{W}_t^W, a)$ for fixed $n$, so $\tilde{s}_t = (\mathcal{W}_t^W, \hat{\rho}_n)$ carries at least as much information as $\mathcal{W}_t^W$ alone. The approximation quality of this proxy state relative to the true belief $\mathbb{P}(s_t \mid H_t)$ is precisely 
\end{remark}

Algorithm~\ref{alg:risk-gated-rl} further introduces the ensemble consisted of $M$ base learners $Q^{(m)}(\tilde{s}, a)$ for $m \in \{1, \dots, M\}$. The local estimation error is $\epsilon_m(\tilde{s}, a) := Q^{(m)}(\tilde{s}, a) - \widetilde{Q}^*(\tilde{s}, a)$, with a cumulative distribution function $F_{\epsilon}(\cdot \mid \tilde{s}, a)$. 

\subsection{Assumptions}
\begin{ass}\label{ass:window_forget}($\gamma$-Exponential Forgetting)
There exist constants $C > 0$ and $\gamma \in (0,1)$ such that for all $t \geq W$:

$$\bigl\|\mathbb{P}(s_t \mid H_t) \;-\; \mathbb{P}(s_t \mid \mathcal{W}_t^W)\bigr\|_{\mathrm{TV}} \;\leq\; C\,\gamma^W$$
where $\|\cdot\|_{\mathrm{TV}}$ denotes total variation distance. This formalises the assumption that the influence of remote history on the current belief decays exponentially in the window length $W$.
\end{ass}

\begin{ass}\label{ass:lipschitz}(Lipschitz Continuity)
The reward function $r$ is $L_r$-Lipschitz in $s$ for every fixed $a$, and the measurement mapping $\omega$ is $L_\omega$-Lipschitz:

$$|r(s, a) - r(s', a)| \leq L_r\,\|s - s'\|, \qquad |\omega(s) - \omega(s')| \leq L_\omega\,\|s - s'\|, \quad \forall\, s, s' \in \mathcal{S}$$

\end{ass}

\begin{ass}\label{ass:duality}(Strong Duality)
The constrained optimisation problem

$$\max_{\pi \in \Pi}\; V^\pi(H_0) \qquad \text{subject to} \quad \rho^\pi(H_t) \leq \tau \;\;\forall\, H_t$$

satisfies strong duality. That is, there exists a finite optimal Lagrange multiplier $\lambda^* \geq 0$ such that:

$$\max_{\pi \in \Pi}\; \min_{\lambda \geq 0}\; \mathcal{L}(\pi, \lambda) \;=\; \min_{\lambda \geq 0}\; \max_{\pi \in \Pi}\; \mathcal{L}(\pi, \lambda)$$

where the Lagrangian is $\mathcal{L}(\pi, \lambda) := V^\pi(H_0) - \lambda\,(\rho^\pi(H_0) - \tau)$. A sufficient condition is the \textbf{Slater condition} where there exists a strictly feasible policy $\pi_0 \in \Pi$ such that $\rho^{\pi_0}(H_t) < \tau$ for all $H_t$.

\end{ass}

\begin{ass}\label{ass:contraction}($\beta$-Contraction)
The discount factor $\beta \in (0,1)$ ensures that both $\mathcal{T}$ and $\widetilde{\mathcal{T}}$ are $\beta$-contractions in the sup-norm:

$$\|\mathcal{T} Q - \mathcal{T} Q'\|_\infty \leq \beta\,\|Q - Q'\|_\infty, \qquad \|\widetilde{\mathcal{T}}\,\widetilde{Q} - \widetilde{\mathcal{T}}\,\widetilde{Q}'\|_\infty \leq \beta\,\|\widetilde{Q} - \widetilde{Q}'\|_\infty$$

By the Banach Fixed-Point Theorem, this guarantees the existence and uniqueness of $Q^*$ and $\widetilde{Q}^*$.
\end{ass}

\begin{ass}\label{ass:obs_noise}(Additive Observation Noise)
The CGM sensor introduces zero-mean additive Gaussian noise, so the observed glucose reading is:

$$o_t \;=\; \omega(s_t) \;+\; \epsilon_t, \qquad \epsilon_t \;\overset{\mathrm{i.i.d.}}{\sim}\; \mathcal{N}(0, \sigma^2), \quad \sigma \geq 0$$

where the noise variables $\{\epsilon_t\}$ are independent of $\{s_t\}$ and of each other. The noiseless setting of Assumption 
\ref{ass:window_forget}, \ref{ass:lipschitz}, \ref{ass:duality} and \ref{ass:contraction} is recovered by setting $\sigma = 0$. 
\end{ass}

\begin{ass}\label{ass:bounded_reward}
The reward function $r(s, a)$ is uniformly bounded such that $|r(s, a)| \le C_{r}$ for all $(s, a) \in \mathcal{S} \times \mathcal{A}$.
\end{ass}

Under this condition, we ensure that $\widetilde{Q}^*$ is bounded and consequently, the error support  $F_{\epsilon}(\cdot \mid \tilde{s}, a)$ is a compact interval $[\epsilon_{\min}, \epsilon_{\max}]$ for all $(s, a) \in \mathcal{S} \times \mathcal{A}$ with $\epsilon_{\min} = \inf \{x : F_{\epsilon}(x \mid \tilde{s}, a) > 0\}$ and $\epsilon_{\max} = \sup \{x : F_{\epsilon}(x \mid \tilde{s}, a) < 1\}$. This boundedness condition aligns with standard Safe RL benchmarks where state and action spaces are strictly constrained by safety boundaries and constraints of physiological systems such as blood glucose limits.

\begin{ass}[Zero-Inclusive Support]\label{ass:zero} $\widetilde{Q}^*$ is contained within the interior of the ensemble's empirical hypothesis support, i.e., $0 \in \text{int}(\text{supp}(F_{\epsilon}))$, which implies $\epsilon_{\min} < 0 < \epsilon_{\max}$. Furthermore, there exists a constant $\eta > 0$ such that for all $(\tilde{s}, a) \in \widetilde{\mathcal{S}} \times \mathcal{A}$:$$ \eta \le F_\epsilon(0 \mid \tilde{s}, a) \le 1 - \eta $$
\end{ass}

\begin{ass}[Conditional i.i.d. Errors\label{ass:conditional_iid} ]Given $(\tilde{s}, a)$, the errors $\{\epsilon_m\}_{m=1}^M$ are independent and identically distributed (i.i.d.) according to $F_{\epsilon}$.\end{ass}

\begin{ass}\label{ass:linear_interpolate}The fixed-point value function $\widetilde{Q}^*$ admits a linear decomposition based on the latent safety states. Specifically, there exist two bounded functions $Q_{\mathrm{safe}}, Q_{\mathrm{fail}} \in \mathcal{Q}$ such that for all $(\tilde{s}, a) \in \widetilde{\mathcal{S}} \times \mathcal{A}$:$$ \widetilde{Q}^*(\tilde{s}, a) = (1 - \hat{\rho}_n(\mathcal{W}_t^W, a)) Q_{\mathrm{safe}}(\tilde{s}, a) + \hat{\rho}_n(\mathcal{W}_t^W, a) Q_{\mathrm{fail}}(\tilde{s}, a) $$
\end{ass}

\begin{ass}[Uniform Concentration]\label{ass:uniform_rho}
For any $\delta \in (0, 1)$, there exists a bound $\epsilon(n, \delta)$ such that:
\[ 
\mathbb{P} \left( \sup_{\tilde{s} \in \widetilde{\mathcal{S}}, a \in \mathcal{A}} \left| \hat{\rho}_n(\mathcal{W}_t^W, a) - \rho(H_t, a) \right| \leq \epsilon(n, \delta) \right) \geq 1 - \delta 
\]
where $\lim_{n \to \infty} \epsilon(n, \delta) = 0$ for any fixed $\delta$.
\end{ass}

\begin{ass}[Optimization Regularity]\label{ass:regularity}
For each proxy state $\tilde{s}_t \in \widetilde{\mathcal{S}}$, the risk-gated decision problem satisfies:
\begin{enumerate}
    \item \textbf{Continuity and Compactness:} The functions $a \mapsto \widetilde{Q}^*(\tilde{s}_t, a)$ and $a \mapsto \hat{\rho}_n(\tilde{s}_t, a)$ are continuous on the compact action space $\mathcal{A}$.
    \item \textbf{Uniqueness:} The unconstrained maximizer $\tilde{\pi}_{\mathrm{POMDP}}(\tilde{s}_t)$ is unique.
    \item \textbf{Strict Feasibility:} There exists an action $a \in \mathcal{A}$ such that $\hat{\rho}_n(\tilde{s}_t, a) < \tau$.
\end{enumerate}
\end{ass}

\subsection{Main Results}
\begin{prop}\label{thm:value_error_bound}
    Under Assumption \ref{ass:window_forget}, \ref{ass:lipschitz}, \ref{ass:duality} and \ref{ass:contraction}, let $Q^*$ be the Oracle Q-function and $\widetilde{Q}^*$ be the proxy Q-function. Let $\lambda^*$ be the optimal Lagrange multiplier from Assumption \ref{ass:duality}, and let $\lambda \geq 0$ be any suboptimal multiplier actually used in $\widetilde{\mathcal{T}}$. For any window length $W \geq 1$ and training sample size $n \geq 1$, the uniform value error satisfies:
$$\|Q^* - \widetilde{Q}^*\|_\infty \;\leq\; \underbrace{\frac{\,L_r L_\omega C\,\gamma^W}{(1-\beta)^2}}_{\displaystyle\text{(I) history truncation}} \;+\; \underbrace{\frac{|\lambda - \lambda^*|}{1-\beta}}_{\displaystyle\text{(II) dual suboptimality}} \;+\; \underbrace{\frac{\lambda^*}{1-\beta}\cdot\mathcal{O}\!\left(\frac{1}{\sqrt{n}}\right)}_{\displaystyle\text{(III) statistical estimation}}$$
\end{prop}

\begin{corollary}\label{cor:bound}
Under the same conditions as Proposition~\ref{thm:value_error_bound}, the proxy state value function $\widetilde{V}^*$ approximates the Oracle state value function $V^*$ with:

$$\|V^* - \widetilde{V}^*\|_\infty \;\leq\; \|Q^* - \widetilde{Q}^*\|_\infty \;\leq\; \frac{\,L_r L_\omega C\,\gamma^W}{(1-\beta)^2} + \frac{|\lambda - \lambda^*|}{1-\beta} + \frac{\lambda^*}{1-\beta}\cdot\mathcal{O}\!\left(\frac{1}{\sqrt{n}}\right)$$
\end{corollary}

\begin{remark}[Error Decomposition]
Proposition~\ref{thm:value_error_bound} bound the value error by three terms; in this remark, we delineate the meaning of each component.
\begin{enumerate}[label=(\Roman*), leftmargin=*, topsep=5pt]
    \item \textbf{History truncation error:} Using a finite window $\mathcal{W}_t^W$ instead of the full history $H_t$ causes value loss. The $(1-\beta)^{-2}$ factor captures two effects: immediate reward approximation error (scaled by $(1-\beta)^{-1}$) and error propagation through future decisions (another $(1-\beta)^{-1}$ factor).
    
    \item \textbf{Dual suboptimality error:} Value loss from using an approximate Lagrange multiplier $\lambda \neq \lambda^*$. This error is always non-negative and equals zero when the constraint is optimally calibrated.
    
    \item \textbf{Statistical estimation error:} Value deviation from replacing the true risk $\rho$ with the learned estimator $\hat{\rho}_n$. The estimation noise $|\rho - \hat{\rho}_n|$ is amplified by the dual multiplier $\lambda^*$ and the effective horizon $(1-\beta)^{-1}$.
\end{enumerate}
\end{remark}
\paragraph{Sketch of proof( Proposition~\ref{thm:value_error_bound}):}
Below are the steps of deriving Proposition~\ref{thm:value_error_bound}. Corollary~\ref{cor:bound} follows directly from Proposition~\ref{thm:value_error_bound} by the non-expansiveness of the maximum operator.

\subparagraph{Step 1: Operator Decomposition via the Banach Fixed-Point Theorem}

We begin by embedding both operators into a common ambient Banach space.
Define the projection map $\Phi: \mathcal{H}_t \to \widetilde{\mathcal{S}}$ by
\[
\Phi(H_t) \;=\; \tilde{s}_t \;=\; \bigl(\mathcal{W}_t^W,\; \hat{\rho}_n(\mathcal{W}_t^W, \cdot)\bigr)
\]
which maps each full history to its proxy state via the local observation window. 
For any $Q \in \mathcal{Q}(\mathcal{H} \times \mathcal{A})$, define its pushforward 
$\bar{Q} \in \mathcal{Q}(\widetilde{\mathcal{S}} \times \mathcal{A})$ by
\[
\bar{Q}(\tilde{s}, a) \;:=\; Q(H_t, a), \qquad \text{where } \Phi(H_t) = \tilde{s}
\]
Under this embedding, both $\mathcal{T}$ and $\widetilde{\mathcal{T}}$ act on the same 
Banach space $\bigl(\mathcal{Q}(\widetilde{\mathcal{S}} \times \mathcal{A}),\, 
\|\cdot\|_\infty\bigr)$, and the triangle inequality is applicable. 
The approximation error introduced by $\Phi$ is precisely the belief bias 
captured by $\varepsilon_1$ and $\varepsilon_3$ in Step~2.

Since $Q^*$ and $\widetilde{Q}^*$ are fixed points of $\mathcal{T}$ and 
$\widetilde{\mathcal{T}}$ respectively, applying the triangle inequality 
in this common space gives:
\[
\|Q^* - \widetilde{Q}^*\|_\infty 
\;=\; \|\mathcal{T} Q^* - \widetilde{\mathcal{T}}\,\widetilde{Q}^*\|_\infty 
\;\leq\; \|\mathcal{T} Q^* - \widetilde{\mathcal{T}} Q^*\|_\infty 
      + \|\widetilde{\mathcal{T}} Q^* - \widetilde{\mathcal{T}}\,\widetilde{Q}^*\|_\infty
\]
The $\beta$-contraction property (Assumption~\ref{ass:contraction}) bounds 
the second term:
\[
\|\widetilde{\mathcal{T}} Q^* - \widetilde{\mathcal{T}}\,\widetilde{Q}^*\|_\infty 
\;\leq\; \beta\,\|Q^* - \widetilde{Q}^*\|_\infty
\]
Rearranging gives the fundamental reduction:
\begin{align}
\|Q^* - \widetilde{Q}^*\|_\infty 
\;\leq\; \frac{1}{1-\beta}\,\|\mathcal{T} Q^* - \widetilde{\mathcal{T}} Q^*\|_\infty
\end{align}

\subparagraph{Step 2: Decomposition of the Operator Gap}
By the triangle inequality, operator gap can be split into three independent error sources:
\begin{align}
    \|\mathcal{T} Q^* - \widetilde{\mathcal{T}} Q^*\|_\infty &\leq \underbrace{\bigl|\mathbb{E}[r \mid H_t, a] - \mathbb{E}_{s \sim p(s|\mathcal{W}_t^W)}[r(s,a)]\bigr|}_{\varepsilon_1:\;\text{belief bias in immediate reward}} \nonumber \\
    &+ \underbrace{\bigl|\lambda^*(\rho - \tau) - \lambda(\hat{\rho}_n - \tau)\bigr|}_{\varepsilon_2:\;\text{dual bias}} \;+\; \underbrace{\beta\,\bigl|\mathbb{E}[V^*(H_{t+1})] - \mathbb{E}[\widetilde{V}^*(\tilde{s}_{t+1})]\bigr|}_{\varepsilon_3:\;\text{belief bias in future value}}
\end{align}

Recall that $V^*(H_t) = \max_{a \in \mathcal{A}_{\mathrm{safe}}^*(H_t)} Q^*(H_t, a)$ is the Oracle state value function.

\textbf{Bounding $\varepsilon_1$ (immediate reward belief bias)}

By the coupling characterization of total variation and Assumption \ref{ass:window_forget} and \ref{ass:lipschitz},
\begin{align}
\varepsilon_1 &= \bigl|\mathbb{E}_{s \sim \mathbb{P}(s \mid H_t)}[r(s,a)] - \mathbb{E}_{s \sim \mathbb{P}(s \mid \mathcal{W}_t^W)}[r(s,a)]\bigr| \nonumber \\
&\leq L_r L_\omega \cdot \|\mathbb{P}(s_t \mid H_t) - \mathbb{P}(s_t \mid \mathcal{W}_t^W)\|_{\mathrm{TV}} \nonumber \\
&\leq L_r L_\omega\, C\,\gamma^W \label{eq:eps1_bound}
\end{align}

The factor $L_\omega$ appears because $r$ depends on $s$ through the observable $\omega(s)$, so the effective Lipschitz constant in the TV metric on the glucose space is $L_r L_\omega$.

\textbf{Bounding $\varepsilon_2$ (dual bias)}

Adding and subtracting $\lambda^*(\hat{\rho}_n - \tau)$:
\begin{equation}
    \varepsilon_2 \;=\; \bigl|\lambda^*(\rho - \tau) - \lambda(\hat{\rho}_n - \tau)\bigr| \;\leq\; |\lambda^* - \lambda|\,|\hat{\rho}_n - \tau| + \lambda^*\,|\rho - \hat{\rho}_n| \;\leq\; |\lambda - \lambda^*| + \lambda^*|\rho - \hat{\rho}_n|
\end{equation}

using $|\hat{\rho}_n - \tau| \leq 1$ since $\hat{\rho}_n \in [0,1]$ and $\tau \in (0,1)$.

\textbf{Bounding $\varepsilon_3$ (future value belief bias)}

The Oracle state value function $V^*(H_t) = \max_{a \in \mathcal{A}_{\mathrm{safe}}^*(H_t)} Q^*(H_t, a)$ inherits $L_r L_\omega$-Lipschitz continuity from $Q^*$ through the Bellman recursion. Applying the same coupling argument to each future step $k = 1, 2, \dots$ and summing the discounted geometric series:

\begin{equation}
    \varepsilon_3 \;\leq\; \sum_{k=1}^{\infty} \beta^k \cdot L_r L_\omega\, C\,\gamma^W \;=\; \frac{\beta}{1-\beta}\cdot L_r L_\omega\, C\,\gamma^W
\end{equation}

Combining $\varepsilon_1$, $\varepsilon_2$, $\varepsilon_3$,

\begin{align}
\|\mathcal{T} Q^* - \widetilde{\mathcal{T}} Q^*\|_\infty &\leq\; L_r L_\omega C\,\gamma^W\!\!\left(1 + \frac{\beta}{1-\beta}\right) + |\lambda - \lambda^*| + \lambda^*|\rho - \hat{\rho}_n| \nonumber \\
&=\; \frac{L_r L_\omega C\,\gamma^W}{1-\beta} + |\lambda - \lambda^*| + \lambda^*|\rho - \hat{\rho}_n|
\end{align}

\subparagraph{Step 3: Statistical Bound via Uniform Concentration}

By Assumption~\ref{ass:uniform_rho} (Uniform Concentration), for any $\delta \in (0,1)$, 
with probability at least $1 - \delta$:
\[
\sup_{\tilde{s} \in \widetilde{\mathcal{S}},\, a \in \mathcal{A}} 
|\hat{\rho}_n(\mathcal{W}_t^W, a) - \rho(H_t, a)| \leq \epsilon(n, \delta)
\]
where $\epsilon(n, \delta) \to 0$ as $n \to \infty$ for any fixed $\delta$.

A canonical instantiation of this bound follows from Hoeffding's inequality. 
Treating the $n$ training labels $y^{(i)} = \mathbf{1}_{\{\omega(s_{t+1}^{(i)}) < g_{\mathrm{low}}\}} \in \{0,1\}$ 
as i.i.d.\ Bernoulli random variables bounded in $[0,1]$, Hoeffding's inequality gives:
\[
\mathbb{P}\!\left(|\rho - \hat{\rho}_n| > \sqrt{\frac{\ln(2/\delta)}{2n}}\right) \leq \delta
\]
which yields $\epsilon(n,\delta) = \sqrt{\tfrac{\ln(2/\delta)}{2n}} = \mathcal{O}(1/\sqrt{n})$ 
for fixed $\delta$. More generally, any concentration inequality 
(e.g., Rademacher complexity bounds or VC-dimension arguments) 
that provides a uniform bound over $(\tilde{s}, a)$ can serve as a valid 
instantiation of Assumption~\ref{ass:uniform_rho}.

Substituting $|\rho - \hat{\rho}_n| \leq \epsilon(n,\delta) = \mathcal{O}(1/\sqrt{n})$ 
into the bound for $\|\mathcal{T} Q^* - \widetilde{\mathcal{T}} Q^*\|_\infty$ and 
combining with Step~1, we obtain with probability at least $1 - \delta$:
\begin{align}
\|Q^* - \widetilde{Q}^*\|_\infty 
&\leq \frac{1}{1-\beta}\!\left[
    \frac{L_r L_\omega C\,\gamma^W}{1-\beta} 
    + |\lambda_{\mathrm{risk}} - \lambda^*| 
    + \lambda^* \cdot \mathcal{O}\!\left(\frac{1}{\sqrt{n}}\right)
\right] \nonumber \\
&= \frac{L_r L_\omega C\,\gamma^W}{(1-\beta)^2} 
+ \frac{|\lambda_{\mathrm{risk}} - \lambda^*|}{1-\beta} 
+ \frac{\lambda^*}{1-\beta} \cdot \mathcal{O}\!\left(\frac{1}{\sqrt{n}}\right)
\end{align}
which completes the proof.

\begin{remark}\label{remark:discussion}
Here lists the discussion and extension of Proposition~\ref{thm:value_error_bound}
\begin{enumerate}[label=(\Roman*), leftmargin=*, nosep]
        \item \textbf{Tightness of the bound:} Term (I) is tight in $W$: as $W \to \infty$, $\gamma^W \to 0$ and the history-truncation error vanishes, recovering the exact Oracle operator. Term (II) vanishes identically when $\lambda = \lambda^*$, i.e., when the proxy operator is calibrated to the true dual optimum. Term (III) decays at the parametric rate $n^{-1/2}$, which is minimax optimal for estimating a Bernoulli probability from i.i.d. observations.
    
    \item \textbf{Data Dependence} While Step 3 adopts an i.i.d. assumption and Assumption~\ref{ass:uniform_rho} for tractability, real-world physiological data exhibits \textit{within-patient dependence} due to temporal correlations in the sliding windows. This structure can be formally addressed by modeling the sequence as a $\beta$-mixing process with geometrically decaying coefficients $\beta_{0k} \leq B\mu^{k},\;\mu \in (0,1)$, reflecting the finite-memory nature of insulin-glucose dynamics. Under this extension, a standard blocking argument \citep{mixing_process} recovers the same concentration rates, albeit with the sample size $n$ replaced by an effective sample size $n_{\text{eff}}$.
    
    \item \textbf{Impact of Observation Noise:} Proposition~\ref{thm:value_error_bound} is derived under the deterministic measurement assumption ($\sigma = 0$). In the presence of CGM measurement noise (Assumption~\ref{ass:obs_noise}), the Lipschitz continuity of the operators ensures that the error components degrade with $\sigma$. We focus on the noiseless case to illustrate the fundamental error decomposition though the extension to $\sigma > 0$ follows similar arguments in our proof.
    \item \textbf{Deep Function Approximation:} Our analysis focuses on the structural error of the proxy MDP itself, assuming the proxy Q-function $\widetilde{Q}^*$ can be exactly represented. In practice, when $\widetilde{Q}^*$ is approximated via deep neural networks, an additional additive term $\mathcal{E}_{\text{approx}}$ appears in the bound, representing the representation capacity and optimization gap. While quantifying $\mathcal{E}_{\text{approx}}$ for general deep architectures remains an open challenge, our bound characterizes the error floor inherent to the risk-gated formulation regardless of the function approximator used.
\end{enumerate}
\end{remark}

\begin{prop}\label{prop:envelope} Under Assumption~\ref{ass:bounded_reward},\ref{ass:zero}, \ref{ass:conditional_iid} and \ref{ass:linear_interpolate}, then for any $(\tilde{s}, a) \in \widetilde{\mathcal{S}} \times \mathcal{A}$, the ensemble safety envelope $[Q^-_M, Q^+_M]$ and the resulting mixed $Q$-value $Q^{\mathrm{gate}}$ satisfy:

\begin{enumerate}

\item \label{lem:enveloping}\textbf{Uniform Exponential Enveloping:} The probability that the fixed point $\widetilde{Q}^*$ is not contained within the ensemble envelope is bounded by:$$ \mathbb{P}\left(\widetilde{Q}^*(\tilde{s}, a) \notin [Q^-_M, Q^+_M]\right) \le (p_+)^M + (p_-)^M\;,$$
where $p_+ = 1-\inf_{(\tilde{s}, a)} F_{\epsilon}(0 \mid \tilde{s}, a)$ and $p_- = \sup_{(\tilde{s}, a)} F_{\epsilon}(0 \mid \tilde{s}, a)$ 

\item \label{lem:consistency}\textbf{Extremal Consistency:} As the ensemble size $M \to \infty$, the envelope boundaries converge in probability to the physical error supports:$$ Q^+_M(\tilde{s}, a) \xrightarrow{p} \widetilde{Q}^*(\tilde{s}, a) + \epsilon_{\max}(\tilde{s}, a) $$$$ Q^-_M(\tilde{s}, a) \xrightarrow{p} \widetilde{Q}^*(\tilde{s}, a) + \epsilon_{\min}(\tilde{s}, a) $$

\item \label{lem:mixed_val}\textbf{Asymptotic Mixed Value:} 
The risk-gated $Q$-value $Q^{\mathrm{gate}}(a)$ converges in probability to a safety-augmented landscape as $M \to \infty$
$$ Q^{\mathrm{gate}}(\tilde{s}, a) \xrightarrow{p} \widetilde{Q}^*(\tilde{s}, a) + \underbrace{(1-\hat{\rho}_n)\epsilon_{\max} + \hat{\rho}_n \epsilon_{\min}}_{\text{safety aware shift}}$$

\end{enumerate}
\end{prop}

\paragraph{Sketch of Proof (Proposition~\ref{prop:envelope}):}
The proof of Proposition~\ref{prop:envelope}-\ref{lem:enveloping} follows from the independence of the $M$ base learners (Assumption~\ref{ass:conditional_iid}) and uniform bounds $p_+, p_-$ from Assumption~\ref{ass:zero}. Proposition~\ref{prop:envelope}-\ref{lem:consistency} is established by noting that $F_{\epsilon}$ has non-vanishing probability mass in every neighborhood of its essential endpoints $\epsilon_{\min}$ and $\epsilon_{\max}$ ( Assumption~\ref{ass:bounded_reward}). Finally, Proposition~\ref{prop:envelope}-\ref{lem:mixed_val} is obtained by substituting the limits of $Q^+_M$ and $Q^-_M$ into the definition of $Q^{\mathrm{gate}}$, utilizing the linearity of the probability limit under the given risk weighting $\hat{\rho}_n$.

\begin{remark}\label{remark:ensemble}
Proposition~\ref{prop:envelope} establishes the theoretical foundation for $Q^{\mathrm{gate}}$ by offering an ensemble envelope guarantee in theory and a safety-aware shift in practice.
\begin{enumerate}[label=(\Roman*), ref=(\Roman*)]\item \textbf{Ensemble envelope guarantee}
Proposition~\ref{prop:envelope}-\ref{lem:enveloping} justifies the convex combination in $Q^{\mathrm{gate}}$ by establishing that $\widetilde{Q}^*$ is strictly contained within the ensemble envelope, while Proposition~\ref{prop:envelope}-\ref{lem:consistency} ensures these boundaries remain well-behaved, converging to the physical limits of the hypothesis space rather than diverging.

\item \textbf{Safety aware shift} Proposition~\ref{prop:envelope}-\ref{lem:mixed_val} justifies $Q^{\mathrm{gate}}$ as a straightforward, risk-sensitive proxy. Although $Q^{\mathrm{gate}}$ introduces a safety-aware shift relative to $\widetilde{Q}^*$, this deviation functions as a \textbf{conservative regularizer} consistent with the Principle of Pessimism \citep{jin2021pessimism}. Similar to Conservative Q-Learning \citep{kumar2020conservative}, this shift enforces a protective margin that prevents value overestimation in high-risk regions. By leveraging extremal statistics to induce this margin, our method bypasses the need for explicit supervision of latent $Q_{\mathrm{safe}}$ and $Q_{\mathrm{fail}}$ components, which are typically unobservable in clinical settings \citep{tamar2016sequential}. This "emergent" safety mechanism aligns with the "Do No Harm" in \citep{gottesman2019guidelines}, ensuring policy robustness under asymmetric risks \citep{bellemare2017distributional}.
\end{enumerate}
\end{remark}

\begin{prop}[Risk-Gated Policy Dynamics]\label{prop:pomdp_agreement}
Under Assumptions~\ref{ass:uniform_rho}--\ref{ass:regularity}, the following hold with probability at least $1 - \delta$ for any history $H_t$:

\begin{enumerate}
    \item \textbf{Policy Agreement:} 
    \[ \mathbb{P} \left( \tilde{\pi}^*(\tilde{s}_t) = \tilde{\pi}_{\mathrm{POMDP}}(\tilde{s}_t) \mid \Delta_{\rho}(\tilde{s}_t) > \epsilon(n, \delta) \right) \geq 1 - \delta \]
    where $\Delta_{\rho}(\tilde{s}_t) := \tau - \hat{\rho}_n(\tilde{s}_t, \tilde{\pi}_{\mathrm{POMDP}}(\tilde{s}_t))$ stands for the empirical safety margin of POMDP.
    \item \textbf{Guaranteed Intervention:} Let $\mathcal{B}_t = \{ \rho(H_t, \tilde{\pi}_{\mathrm{POMDP}}(\tilde{s}_t)) > \tau + \epsilon \}$ be the event where the POMDP policy is truly unsafe. Then:
    \[ \mathbb{P} \left( \tilde{\pi}^* \neq \tilde{\pi}_{\mathrm{POMDP}} \text{ and } \rho(H_t, \tilde{\pi}^*) \leq \tau + \epsilon \mid \mathcal{B}_t \right) \geq 1 - \delta \]
\end{enumerate}
\end{prop}
\paragraph{Sketch of proof (Proposition~\ref{prop:pomdp_agreement}):}
Consider $$ \mathcal{E}_n := \left\{ \omega \in \Omega : \sup_{\tilde{s}, a} \left| \hat{\rho}_n(\tilde{s}, a) - \rho(H_t, a) \right| \leq \epsilon(n, \delta) \right\} $$By Assumption~\ref{ass:uniform_rho}, we have $\mathbb{P}(\mathcal{E}_n) \geq 1-\delta$. All subsequent arguments hold deterministically given $\mathcal{E}_n$.

\subparagraph{Part 1: Policy Agreement}
On the event $\mathcal{E}_n$, suppose the empirical safety margin satisfies $\Delta_{\rho}(\tilde{s}_t) > \epsilon(n, \delta)$. 
By substituting the definition of $\Delta_{\rho}$, we obtain:$$ \tau - \hat{\rho}_n(\tilde{s}_t, \tilde{\pi}_{\mathrm{POMDP}}) > \epsilon(n, \delta) \implies \hat{\rho}_n(\tilde{s}_t, \tilde{\pi}_{\mathrm{POMDP}}) < \tau - \epsilon(n, \delta) $$

This ensures $\hat{\rho}_n(\tilde{s}_t, \tilde{\pi}_{\mathrm{POMDP}}) < \tau$, placing the unconstrained POMDP policy within the empirical safe set $\mathcal{A}_{\mathrm{safe}}(\tilde{s}_t)$. Since $\tilde{\pi}_{\mathrm{POMDP}}$ is the global maximizer of $\widetilde{Q}^*$ over the entire action space (Assumption~\ref{ass:regularity}), it must also be the maximizer when restricted to $\mathcal{A}_{\mathrm{safe}}$, yielding $\tilde{\pi}^*(\tilde{s}_t) = \tilde{\pi}_{\mathrm{POMDP}}(\tilde{s}_t)$.

\subparagraph{Part 2: Guaranteed Intervention}
We consider the conditional event $\mathcal{B}_t \cap \mathcal{E}_n$. Given $\rho(H_t, \tilde{\pi}_{\mathrm{POMDP}}) > \tau + \epsilon(n, \delta)$, the uniform concentration on $\mathcal{E}_n$ implies:
$$ \hat{\rho}_n(\tilde{s}_t, \tilde{\pi}_{\mathrm{POMDP}}) \geq \rho(H_t, \tilde{\pi}_{\mathrm{POMDP}}) - \epsilon(n, \delta) > (\tau + \epsilon) - \epsilon = \tau .$$
This ensures $\tilde{\pi}_{\mathrm{POMDP}} \notin \mathcal{A}_{\mathrm{safe}}$, forcing the Risk-Gated policy to diverge: $\tilde{\pi}^* \neq \tilde{\pi}_{\mathrm{POMDP}}$. 
Simultaneously, our algorithm guarantees $\hat{\rho}_n(\tilde{s}_t, \tilde{\pi}^*) \leq \tau$ (Assumption~\ref{ass:regularity}). Invoking $\mathcal{E}_n$ again to bound the population risk:$$ \rho(H_t, \tilde{\pi}^*) \leq \hat{\rho}_n(\tilde{s}_t, \tilde{\pi}^*) + \epsilon(n, \delta) \leq \tau + \epsilon(n, \delta) $$Thus, on the high-probability event $\mathcal{E}_n$, both the intervention and the safety bound hold.
\begin{remark}[Behavioral Implications of Uncertainty]\label{remark:POMDP}
Proposition~\ref{prop:pomdp_agreement} characterizes the operational regions of the Risk-Gated policy; we delineate the implications of the statistical bound $\epsilon(n, \delta)$ as follows:
\begin{enumerate}[label=(\Roman*), leftmargin=*, topsep=5pt]
   \item \textbf{Statistical Confidence as a Gate:} The condition $\Delta_{\rho} > \epsilon(n, \delta)$ functions as a \textbf{statistical test}. By requiring the empirical margin to exceed the error bound, the gate ensures that policy agreement only occurs when the evidence for safety is strong enough to overwhelm the estimation noise.
    
    \item \textbf{Calibration of the Safety Threshold:} The presence of $\epsilon$ in the intervention bound $\tau + \epsilon$ indicates that the "true" safety limit of the system is not just the threshold $\tau$, but an \textbf{inflated threshold} adjusted for epistemic uncertainty. This shows that safety is not a static point but a function of the data density $n$.
    
    \item \textbf{Asymptotic Alignment:} As the sample size $n \to \infty$, the error bound $\epsilon(n, \delta) \to 0$, causing the "gray zone" to vanish. In this limit, the Risk-Gated policy perfectly aligns with the POMDP whenever it is safe ($\rho \leq \tau$) and intervenes precisely when it is not ($\rho > \tau$), achieving \textbf{statistical consistency}.
\end{enumerate}
\end{remark}

\section{Additional Experimental Details}
\label{app:experiment_details}

\paragraph{Glucose-control details.}
For each virtual patient, we use a multi-day meal scenario with four daily meals scheduled at approximately 08:00, 12:30, 16:00, and 19:00, with carbohydrate amounts specified in the simulator metadata. In the glucose-control instantiation, the proxy state includes current CGM, CGM trend, previous bolus, and a decayed insulin-on-board summary. To stabilize early training and reduce unsafe exploration, we use a rule-based teacher policy during an initial warmup period. The teacher's actions are projected onto the discrete action grid, and the resulting transitions are retained for subsequent learning. In addition to learned risk scoring, hard safety shields suppress insulin delivery in clearly unsafe low-glucose or downward-trending regimes.

\paragraph{Safety-Gym details.}
For Safety-Gym, the proxy state is constructed from recent observations, recent actions, and short-horizon safety-relevant interaction summaries. The controller predicts action-conditioned near-term hazard, combines this signal with ensemble-based value estimates, and selects actions through risk-gated evaluation using the same high-level algorithm as in glucose control.

\subsection{Compute Resources}
\label{app:compute_resources}

All experiments were run by a single worker on a MacBook Pro with an Apple M2 CPU and 16GB unified memory. No external GPU or compute cluster was used. For glucose-control experiments, training and evaluation runtimes are reported in Table~\ref{tab:glucose_results} under the same local CPU-only setup. Safety-Gym experiments were also run on the same machine using the CPU-only configuration.

\subsection{Random Seeds and Reproducibility}
\label{app:random_seeds}

Unless otherwise specified, experiments with repeated runs were evaluated over three random seeds: 42, 123, and 456. These seeds were used for model initialization, environment stochasticity, and sampling where applicable. Reported error bars correspond to mean $\pm$ standard deviation across these seeds.

\section{Domain-Specific Instantiations of the Surrogate View}
\label{app:domain_instantiations_appendix}

The surrogate interpretation is useful precisely because it carries across different partially observable safety-critical domains.

\subsection{Glucose regulation}

In glucose control, the latent state includes hidden physiological quantities such as meal absorption, delayed insulin action, and internal metabolic variation. The controller does not observe these variables directly. Instead, it receives CGM measurements and recent actuation history, from which it constructs a proxy state summarizing current glucose, short-term trend, prior bolus delivery, and insulin-on-board-like information. For each candidate bolus action, it then predicts a near-term hazard estimate reflecting the chance of moving toward unsafe glucose levels. In this domain, the main practical advantage of the surrogate view is that it avoids explicit physiological belief tracking and online planning over hidden meal and insulin states.

\subsection{Safety-constrained navigation}

In Safety-Gym, the latent state is less physiological but still partially observable in the sense that the agent must act under incomplete local information and uncertain future safety consequences. The controller builds a proxy state from recent observations and recent actions and predicts an action-conditioned near-term hazard indicating the likelihood of entering unsafe regions or incurring safety cost shortly after the candidate action. The same mixed-value rule is then used to interpolate between optimistic and conservative action evaluations. Thus, even though the domain differs substantially from glucose regulation, the surrogate interpretation remains the same: the controller acts from compressed recent context and explicit predicted short-horizon hazard rather than from full latent-state inference.

\section{Scope and limitations of the appendix interpretation}
\label{app:appendix_limitations}

The constructions in this appendix are intended only as interpretive support for the main method. They do not constitute a proof that the pair \((o_t,\hat\rho_t(a))\) is a sufficient statistic for optimal control, nor do they establish a formal equivalence between the proposed algorithm and an exact surrogate MDP solution. In particular, a scalar or low-dimensional near-term risk signal cannot generally encode all information in a full belief state.

Instead, the purpose of the appendix is narrower: to clarify why predictive near-term risk may be a useful substitute for explicit belief-state inference in safety-critical problems where short-horizon hazard dominates the practical decision burden. This is the sense in which the method should be understood as a decision-oriented approximation to belief-based control rather than a complete replacement for POMDP reasoning in general.


\end{document}

%% file: math_commands.tex
\newtheorem{corollary}{Corollary}
\newtheorem{prop}{Proposition}
\newtheorem{ass}{Assumption}
\newtheorem{remark}{Remark}

%% file: wip.tex
\section{Risk-Gated Control under Partial Observability}

\subsection{Problem Setting and History-Based Proxy State}

We formulate risk-gated control as a decision-oriented surrogate for safety-critical POMDPs. At time \(t\), the environment has an unobserved latent state \(s_t\in\mathcal S\), while the controller receives a partial observation \(y_t\in\mathcal O\). It selects an action \(a_t\in\mathcal A\), receives task reward \(r_t\), and observes a safety cost or violation signal \(c_t\). Let
\begin{equation}
H_t=(y_1,a_1,r_1,c_1,\ldots,y_{t-1},a_{t-1},r_{t-1},c_{t-1},y_t)    
\end{equation}
denote the available history. In a classical POMDP, the sufficient statistic for optimal control is the belief state $b_t=\mathbb{P}(s_t\mid H_t)$. Instead of maintaining and planning in belief space
$b_t$ over time, 
we compress the most recent $W$ steps of history into a proxy state 
\begin{equation}
o_t=\phi(\mathcal{W}_t^W)\in\widetilde{\mathcal S},
\end{equation}
where $\mathcal{W}_t^W=(y_{t-W+1},a_{t-W+1},\ldots,y_t)$ and $\phi(\cdot)$ aggregates recent observations, actions, and delayed actuation effects. This proxy state serves as the information input for both risk prediction and value learning. The controller therefore avoids explicit belief updates and online belief-space planning, while adapting its conservatism to the predicted hazard of each candidate action.

This proxy state serves as the information input for both risk prediction and value learning. The controller therefore avoids explicit belief updates and online belief-space planning, while adapting its conservatism to the predicted hazard of each candidate action.

\subsection{Action-Conditioned Risk and Risk-Gated Value Estimation}
\label{sec:risk_value}

Given the proxy state $o_t=\phi(\mathcal{W}_t^W)$, the controller evaluates each candidate action $a\in\mathcal A$ using an action-conditioned near-term risk predictor
\begin{equation}
    \hat{\rho}(o_t,a)\in[0,1].
\end{equation}
Here, $\hat{\rho}(o_t,a)$ estimates the likelihood or severity of entering an unsafe region over a short future horizon after taking action $a$. The key distinction from state-only safety estimates is that the risk is computed separately for each candidate action. Thus, under the same partial observation, the controller can distinguish actions that are likely safe from those that may induce near-term hazards. In glucose regulation, for example, different bolus doses can have substantially different near-term hypoglycemia risk even when the current CGM value is identical; in Safety-Gym, different movement actions can create different collision risks from the same local observation.

To use this risk signal for value-based control, we maintain an ensemble of $M$ critics $\{Q_m(o,a)\}_{m=1}^{M}$
The ensemble provides both an optimistic and a conservative estimate of the action value:
$Q^{+}(o,a)=\max_{m} Q_m(o,a), \; Q^{-}(o,a)=\min_{m} Q_m(o,a)$. Rather than using either estimate alone, we interpolate between them according to the predicted risk:
\begin{equation}
    Q^{\mathrm{gate}}(o,a)
    =
    (1-\hat{\rho}(o,a))Q^{+}(o,a)
    +
    \hat{\rho}(o,a)Q^{-}(o,a).
    \label{eq:qmix}
\end{equation}
When $\hat{\rho}(o,a)$ is small, the gated value is close to the optimistic envelope $Q^{+}$, encouraging reward-seeking behavior in low-risk regions. When $\hat{\rho}(o,a)$ is large, the gated value shifts toward the conservative envelope $Q^{-}$, discouraging actions whose estimated returns are uncertain or potentially unsafe. In this way, $Q^{\mathrm{gate}}$ acts as a soft risk-aware value adjustment before the final action filter is applied.

\subsection{Policy Learning and Action Selection}
\label{sec:policy_learning}

The controller combines the soft risk-aware value estimate in Eq.~\eqref{eq:qmix} with a hard admissibility filter. At decision time, it first evaluates the predicted risk of each candidate action and constructs the safe action set
\begin{equation}
    \mathcal A_{\mathrm{safe}}(o_t)
    =
    \{a\in\mathcal A:\hat{\rho}(o_t,a)\leq R_{\max}\},
    \label{eq:safe_set}
\end{equation}
where $R_{\max}$ is the user-specified risk threshold. If this set is nonempty, the controller selects the admissible action with the largest risk-gated value:
\begin{equation}
    a_t
    =
    \arg\max_{a\in\mathcal A_{\mathrm{safe}}(o_t)}
    Q^{\mathrm{gate}}(o_t,a).
    \label{eq:safe_action}
\end{equation}
If no action satisfies the threshold, the controller does not maximize value over unsafe actions. Instead, it falls back to the action with minimum predicted risk:
\begin{equation}
    a_t
    =
    \arg\min_{a\in\mathcal A}
    \hat{\rho}(o_t,a).
    \label{eq:fallback_action}
\end{equation}
This fallback rule ensures that the controller remains well-defined even when all candidate actions are estimated to be risky.

The risk predictor and critic ensemble are trained jointly from replay data. Each transition contains the proxy state, selected action, reward, safety signal, and next observation. The risk model is trained to predict near-term safety violations or pulse-based hazard labels from $(o_t,a_t)$, while the critics are trained using temporal-difference learning. In practice, the reward used for critic learning can include a risk-aware penalty, so that the value function reflects both task performance and safety burden. The hard action filter in Eq.~\eqref{eq:safe_set} is then applied at decision time, making the learned policy both value-seeking and risk-constrained.

\subsection{Connection to Foundational RL Frameworks}
\label{sec:connection}
How does Risk-Gated RL sit within the existing RL landscape? We address this by examining its properties from three perspectives: 
alignment with Safe RL, policy agreement with POMDPs, and the 
safety-aware motivation for $Q^{\mathrm{gate}}$. To clarify these 
links, we distinguish two theoretical notions:
(i) the Safe RL Oracle ($Q^{\star}$), defined over full state history 
with the true risk $\rho$; and
(ii) the Risk-Gated Proxy ($\widetilde{Q}^{\mathrm{gate}}$), which 
operates on windowed proxy states with the empirical risk $\hat{\rho}$.
The practically implemented $Q^{\mathrm{gate}}$ is a conservative 
refinement of $\widetilde{Q}^{\mathrm{gate}}$, as detailed in the 
third perspective below.
This section focuses on informal statements of the core properties; 
full technical results are deferred to Appendix~\ref{sec:theory}.

\paragraph{Alignment with Safe RL}
Given the shared objective of constrained optimization, we first 
benchmark $\widetilde{Q}^{\mathrm{gate}}$ against the Safe RL Oracle 
$Q^{\star}$. The discrepancy arises from three sources: approximation 
via windowed proxy states, penalty mismatch, and risk estimation error. 
Informally,
\begin{equation}\label{eq:safe_rl_bound}
\left\|Q^\star - \widetilde{Q}^{\mathrm{gate}}\right\|_\infty
\;\leq\;
\frac{\epsilon_{\mathrm{proxy}}(W)}{(1-\beta)^2}
+\frac{\epsilon_{\lambda}}{1-\beta}
+\frac{\epsilon_{\rho}}{1-\beta},
\end{equation}
where $\beta\in(0,1)$ is the discount factor; $\epsilon_{\mathrm{proxy}}(W)$ 
captures information loss from the windowed proxy relative to full history; 
$\epsilon_{\lambda}$ denotes the mismatch between the safe-RL penalty and 
our gate-induced penalty; and $\epsilon_{\rho}$ is the estimation error of 
$\hat{\rho}$. This bound confirms that $\widetilde{Q}^{\mathrm{gate}}$ 
recovers the oracle as $W\to\infty$ and $n\to\infty$; see 
Proposition~\ref{thm:value_error_bound} and Remark~\ref{remark:discussion}.
To validate Eq.~\eqref{eq:safe_rl_bound} empirically, 
Section~\ref{sec:risk_predict} evaluates risk estimation performance, 
while Section~\ref{sec:sensitivity} ablates sensitivity to $W$ and 
$\lambda_\mathrm{risk}$.

\paragraph{Policy agreement with POMDP}
Having established the gap to the safe RL oracle, we next ask whether 
safety comes at the cost of reward performance. Our framework provides 
a reassuring answer: by focusing exclusively on near-term risk rather 
than maintaining a full belief state, we avoid the computational burden 
of belief tracking while preserving reward performance. When safety 
margins are satisfied, $\tilde{\pi}^{\mathrm{gate}}$ agrees with 
$\tilde{\pi}_{\mathrm{POMDP}}$ with high probability. Conversely, if 
the safety margin is breached, the framework intervenes to restore the 
safe regime. See Proposition~\ref{prop:pomdp_agreement} and 
Remark~\ref{remark:POMDP} for formal statements and discussion.
\paragraph{Motivation for $Q^{\mathrm{gate}}$}
Beyond policy agreement, we motivate the specific form of 
$Q^{\mathrm{gate}}$ from a conservative decision perspective. While 
$\widetilde{Q}^{\mathrm{gate}}$ achieves theoretical optimality, 
practical deployment in safety-critical domains demands a more 
conservative stance. Grounded in our theoretical guarantee 
(Proposition~\ref{prop:envelope}) that $\widetilde{Q}^{\mathrm{gate}}$ 
lies within $[Q^-, Q^+]$ with high probability, and inspired by the 
Principle of Pessimism~\citep{jin2021pessimism}, we implement 
$Q^{\mathrm{gate}}$ as a convex combination of these bounds weighted 
by $\hat{\rho}$, naturally shifting toward conservatism as risk 
increases in the spirit of Conservative Q Learning~\citep{kumar2020conservative} 
and related frameworks~\citep{bellemare2017distributional,
gottesman2019guidelines,tamar2016sequential}. The precise relationship 
between $Q^{\mathrm{gate}}$ and $\widetilde{Q}^{\mathrm{gate}}$ is 
formalized in Remark~\ref{remark:ensemble}.

These perspectives confirm that Risk-Gated RL provides safety 
guarantees, belief-free tractability, and pragmatic caution. Collectively, 
they yield the competitive risk-adjusted rewards demonstrated in 
Section~\ref{sec:experiment}.

\begin{algorithm}[ht]
\caption{Risk-Gated Reinforcement Learning}
\label{alg:risk-gated-rl}
\small
\begin{algorithmic}[1]
\REQUIRE Action set $\mathcal{A}$, risk threshold $R_{\max}$, risk penalty $\lambda_{\mathrm{risk}}$, ensemble critics $\{Q^{(m)}\}_{m=1}^M$
\FOR{each environment step $t$}
    \STATE Construct proxy state $o_t=\phi(\mathcal{W}_t^W)$ from the $W$ most recent observations and actions.
    \STATE For each $a\in\mathcal{A}$, estimate near-term risk $\hat\rho_t(a)$.
    \STATE Compute \(Q^+(a)=\max_m Q^{(m)}(o_t,a,\hat\rho_t(a))\) and \(Q^-(a)=\min_m Q^{(m)}(o_t,a,\hat\rho_t(a))\).
    \STATE Compute \(Q^{\mathrm{gate}}(a)=(1-\hat\rho_t(a))Q^+(a)+\hat\rho_t(a)Q^-(a)\).
    \STATE Select the admissible action with largest \(Q^{\mathrm{gate}}\), or the minimum-risk action if none is admissible.
    \STATE Execute \(a_t\), store \(\tilde r_t=r_t-\lambda_{\mathrm{risk}}\hat\rho_t(a_t)\), and update the risk model and critics.
\ENDFOR
\end{algorithmic}
\end{algorithm}

%% file: experiments.tex
\section{Experiments}\label{sec:experiment}

\subsection{Experimental Setup}

We evaluate the proposed framework in four settings across two safety-critical control domains: adult glucose regulation, adolescent glucose regulation, SafetyPointGoal1, and SafetyPointCircle1. Additional implementation details are provided in Appendix~\ref{app:experiment_details}.

\paragraph{Glucose control.}
We evaluate glucose regulation in the UVa/Padova simulator \cite{FDA_simulator} through the open-source \texttt{simglucose} package \cite{simglucose_code,zhou2022design}. The controller receives CGM measurements and selects discretized bolus actions every 3 minutes under partial observability. For each virtual patient, we train from April 19, 2025 to May 19, 2025 and evaluate for the following 14 days without resetting the learned model. We report results separately for adult and adolescent cohorts.

\paragraph{Safety-Gym navigation.}
We evaluate safety-constrained navigation on \texttt{SafetyPointGoal1} and \texttt{SafetyPointCircle1} in \texttt{Safety Gym} \cite{safety_gym}. These tasks require the agent to balance task reward against safety cost under limited local observations and uncertain near-term safety consequences. The same risk-gated controller is used, with domain-specific proxy features and hazard estimation adapted to the navigation setting.

\paragraph{Baselines and metrics.}
For glucose regulation, we compare against POMDP \cite{pomdp}, PID \cite{cameron2011riskmanagement}, Meal-Bolus \cite{herrero2017adaptivebolus}, PPO \cite{ppo}, Safe-PPO \cite{zhao2025safeppo}, and Safe-DQN-style glucose-control baselines \cite{tejedor2023dqn}. For Safety-Gym, we compare against PPO, PPO-Lag and TRPO-Lag \cite{ppo-lag,trpo}, CPPO-PID \cite{cppo-pid}, RCPO \cite{rcpo}, CPO \cite{cpo}, PCPO \cite{pcpo}, CUP \cite{cup}, FOCOPS \cite{focops}, and POMDP. Glucose-control metrics include time in range (TIR), time below 70 mg/dL, time above 180 mg/dL, mean blood glucose, and runtime. Safety-Gym metrics include average reward, average safety cost, normalized reward $J^R$, normalized cost $J^C$, and reward-per-cost. 




\subsection{Main Results}

\paragraph{Glucose Control.} We first evaluate glucose regulation, where partial observability stems from unobserved meals, delayed insulin action, and latent physiological variability. 

\begin{figure*}[ht]
    \centering
    \begin{minipage}[ht]{0.45\textwidth}
        \centering
        \includegraphics[width=\linewidth]{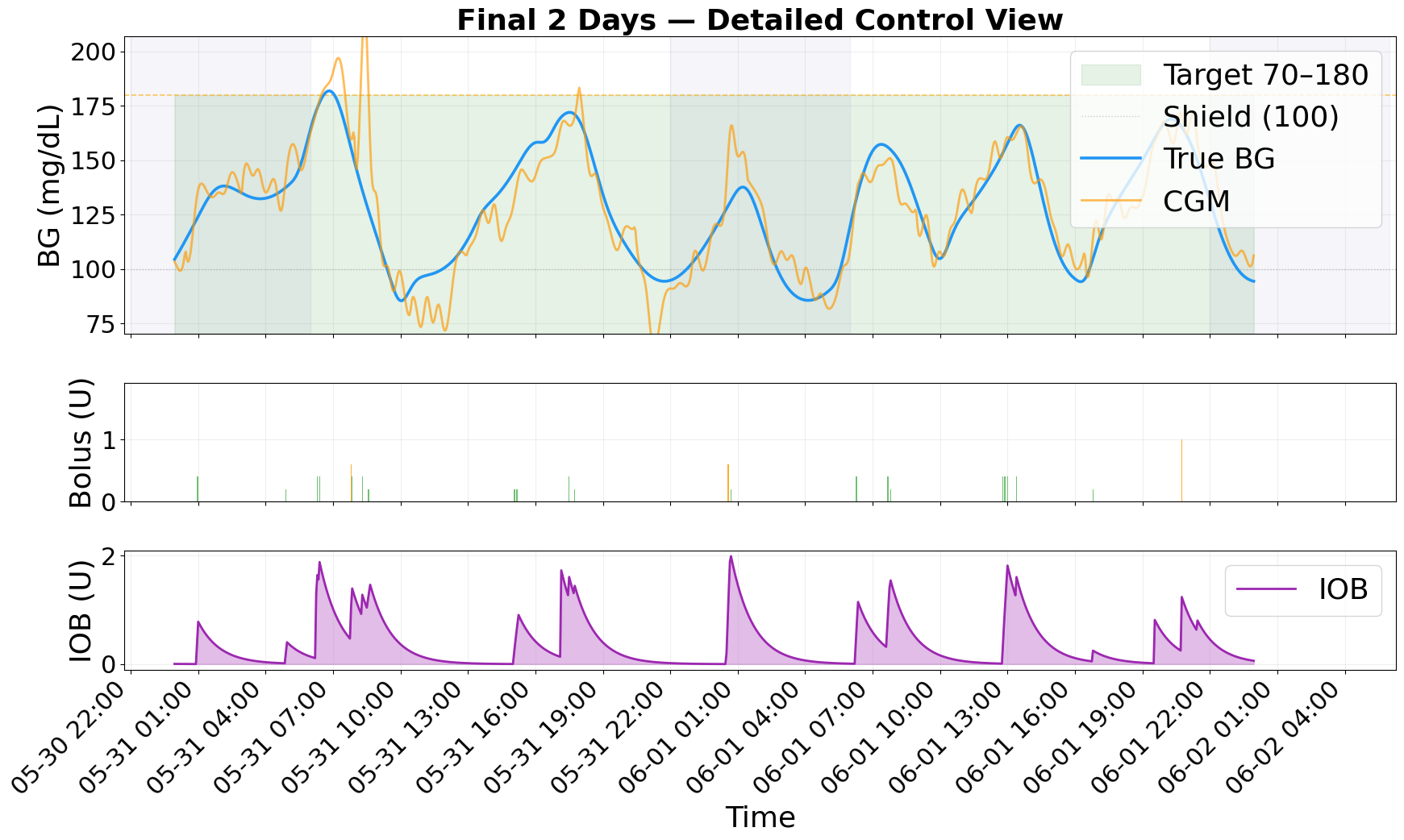}
        \vspace{2mm}
        \small (a) RiskGated
    \end{minipage}\hfill
    \begin{minipage}[ht]{0.45\textwidth}
        \centering
        \includegraphics[width=\linewidth]{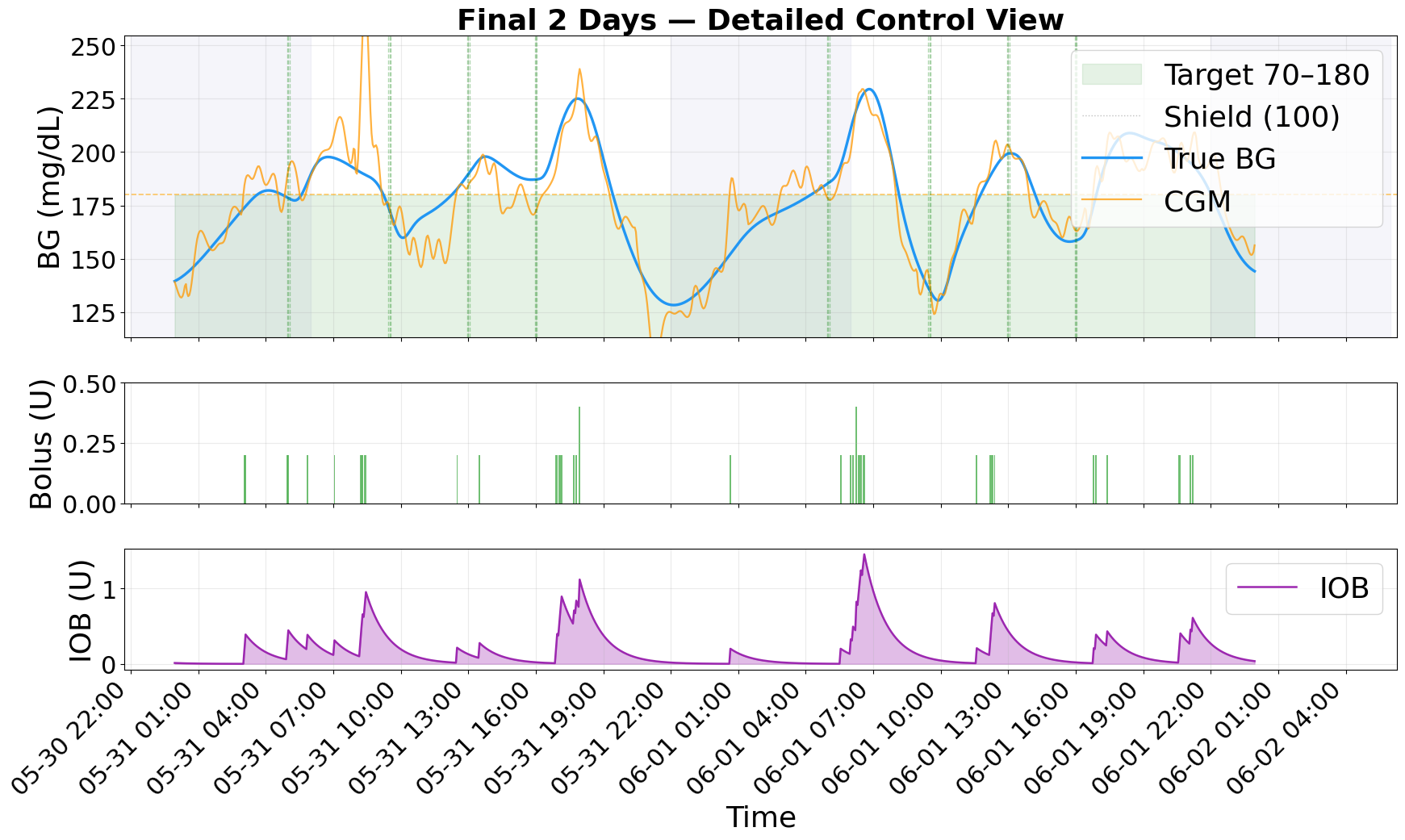}
        \vspace{2mm}
        \small (b) POMDP
    \end{minipage}
    \caption{Qualitative comparison of control behavior over the final 2 evaluation days.}
    \label{fig:qualitative_compare}
\end{figure*}

Table~\ref{tab:glucose_results} reports cohort-level glucose-control results. RiskGated achieves the highest TIR in both cohorts: 82.0\% for adults and 71.6\% for adolescents. Compared with PPO and POMDP, it improves TIR by 4.7\% and 7.6\% on adults, and by 5.9\% and 14.7\% on adolescents, respectively. Although RiskGated is not the best on every individual metric, the competing methods reveal clear tradeoffs: Safe-PPO and Safe-DQN reduce hypoglycemia in adults but incur substantially higher hyperglycemia, while POMDP reduces adolescent hypoglycemia but produces much higher time above range. RiskGated achieves the best mean rank in both cohorts, indicating the strongest aggregate balance across TIR, hypoglycemia, and hyperglycemia. It also substantially reduces computation relative to POMDP planning, lowering total runtime from 1351\,s to 144\,s on adults and from 1346\,s to 136\,s on adolescents, with per-step latency reduced from 61.914\,ms to 4.363\,ms.
The qualitative trajectories in Figure~\ref{fig:qualitative_compare} help explain this tradeoff. RiskGated delivers insulin at fewer, more targeted decision points than Pure POMDP, consistent with risk-gated selection: low predicted risk avoids unnecessary intervention, while elevated risk induces more conservative behavior.

\begin{table}[ht]
\centering
\caption{Glucose-control results over 14 days on 10 virtual patients. Values are mean $\pm$ standard deviation. Mean Rank averages ordinal ranks over TIR, Time$<70$, and Time$>180$.}
\label{tab:glucose_results}
\setlength{\tabcolsep}{4pt}
\resizebox{\columnwidth}{!}{%
\begin{tabular}{lccc|cc|c}
\toprule
\textbf{Algorithm} 
& \multicolumn{3}{c|}{\textbf{Primary glucose metrics}}
& \multicolumn{2}{c|}{\textbf{Secondary metrics}}
& \textbf{Summary} \\
\cmidrule(r){2-4}\cmidrule(r){5-6}\cmidrule(l){7-7}
& \textbf{TIR (\%) $\uparrow$}
& \textbf{Time$<70$ (\%) $\downarrow$}
& \textbf{Time$>180$ (\%) $\downarrow$}
& \textbf{Mean BG}
& \textbf{Runtime (s) $\downarrow$}
& \textbf{Mean Rank $\downarrow$} \\
\midrule
\multicolumn{7}{l}{\textbf{Adult}} \\
PID          & 80.4 $\pm$ 18.8 & 9.7 $\pm$ 13.3 & 10.0 $\pm$ 7.2 & 125.6 $\pm$ 9.8 & 81 & 2.67 \\
Meal-Bolus   & 58.4 $\pm$ 9.2  & 26.3 $\pm$ 8.6 & 15.3 $\pm$ 11.0 & 117.7 $\pm$ 14.8 & 72 & 6.00 \\
PPO          & 77.3 $\pm$ 16.9 & 2.2 $\pm$ 4.5 & 20.5 $\pm$ 17.2 & 144.0 $\pm$ 17.2 & 83 & 3.67 \\
Safe-PPO     & 76.0 $\pm$ 18.5 & 1.9 $\pm$ 3.9 & 22.1 $\pm$ 18.8 & 145.9 $\pm$ 18.8 & 77 & 3.83 \\
Safe-DQN     & 75.7 $\pm$ 16.1 & 1.9 $\pm$ 4.5 & 22.4 $\pm$ 16.1 & 146.2 $\pm$ 16.0 & 78 & 4.50 \\
POMDP        & 74.4 $\pm$ 16.5 & 11.6 $\pm$ 12.2 & 14.0 $\pm$ 8.1 & 129.1 $\pm$ 12.2 & 1351 & 5.00 \\
\textbf{RiskGated} & $\mathbf{82.0 \pm 12.0}$ & 4.2 $\pm$ 7.2 & 13.8 $\pm$ 10.6 & 135.8 $\pm$ 16.9 & 144 & $\mathbf{2.33}$ \\

\midrule
\multicolumn{7}{l}{\textbf{Adolescent}} \\
PID             & 66.6 $\pm$ 21.0 & 20.3 $\pm$ 15.4 & 13.1 $\pm$ 10.7 & 117.1 $\pm$ 14.0 & 78 & 3.00 \\
Meal-Bolus      & 54.8 $\pm$ 13.7 & 21.3 $\pm$ 15.5 & 24.0 $\pm$ 9.2 & 130.9 $\pm$ 19.3 & 70 & 6.67 \\
PPO             & 65.7 $\pm$ 17.1 & 13.8 $\pm$ 14.4 & 20.5 $\pm$ 18.8 & 132.3 $\pm$ 30.3 & 77 & 4.33 \\
Safe-PPO        & 66.3 $\pm$ 17.4 & 13.3 $\pm$ 14.2 & 20.4 $\pm$ 19.5 & 132.7 $\pm$ 31.1 & 74 & 3.33 \\
Safe-DQN+IOB    & 65.6 $\pm$ 17.5 & 15.1 $\pm$ 14.7 & 19.3 $\pm$ 19.2 & 130.3 $\pm$ 32.6 & 74 & 4.33 \\
POMDP           & 56.9 $\pm$ 8.6  & 10.0 $\pm$ 8.0 & 33.1 $\pm$ 7.7 & 158.6 $\pm$ 16.0 & 1346 & 4.67 \\
\textbf{RiskGated} & $\mathbf{71.6 \pm 16.5}$ & 11.2 $\pm$ 11.1 & 17.2 $\pm$ 14.8 & 129.4 $\pm$ 21.5 & 136 & $\mathbf{1.67}$ \\
\bottomrule
\end{tabular}%
}
\end{table}

\begin{figure*}[ht]
\centering

\begin{minipage}[t]{0.56\textwidth}
\centering
\captionof{table}{Safety-Gym results on SafetyPointGoal1 and SafetyPointCircle1. Values are mean $\pm$ standard deviation. Higher reward, $J^R$, and Reward/Cost are better; lower cost and $J^C$ are better.}
\label{tab:safety_benchmark}
\setlength{\tabcolsep}{2.5pt}
\scriptsize
\resizebox{\linewidth}{!}{%
\begin{tabular}{l llll c}
\toprule
\textbf{Algorithm} & \textbf{AvgReward $\uparrow$} & \textbf{AvgCost $\downarrow$} & \textbf{$J^R$ $\uparrow$} & \textbf{$J^C$ $\downarrow$} & \textbf{Reward/Cost $\uparrow$} \\
\midrule
\multicolumn{6}{l}{\textbf{SafetyPointGoal1}} \\
PPO & 11.37 $\pm$ 1.39 & 109.87 $\pm$ 43.86 & 1.000 & 4.395 & 0.103 \\
PPO-Lag & 8.14 $\pm$ 3.52 & 90.27 $\pm$ 34.98 & 0.866 & 3.611 & 0.090 \\
TRPO-Lag & 2.99 $\pm$ 2.06 & 195.33 $\pm$ 215.45 & 0.318 & 7.813 & 0.015 \\
CPPO-PID & 0.68 $\pm$ 0.71 & 110.67 $\pm$ 156.51 & 0.072 & 4.427 & 0.006 \\
RCPO & 2.99 $\pm$ 2.06 & 195.33 $\pm$ 215.45 & 0.318 & 7.813 & 0.015 \\
CPO & -0.12 $\pm$ 0.46 & 25.67 $\pm$ 36.30 & -0.013 & 1.027 & -0.005 \\
PCPO & -0.73 $\pm$ 1.19 & 5.67 $\pm$ 8.01 & -0.077 & 0.227 & -0.129 \\
CUP & 2.30 $\pm$ 2.62 & 41.67 $\pm$ 23.41 & 0.245 & 1.667 & 0.055 \\
FOCOPS & 2.26 $\pm$ 2.51 & 309.33 $\pm$ 388.72 & 0.241 & 12.373 & 0.007 \\
POMDP & -0.71 $\pm$ 1.95 & 101.50 $\pm$ 26.33 & -0.063 & 4.060 & -0.007 \\
\textbf{RiskGated} & 5.39 $\pm$ 2.35 & 36.93 $\pm$ 10.32 & 0.474 & 1.477 & \textbf{0.146} \\
\midrule
\multicolumn{6}{l}{\textbf{SafetyPointCircle1}} \\
PPO & 25.15 $\pm$ 9.46 & 278.48 $\pm$ 32.10 & 1.000 & 11.139 & 0.090 \\
PPO-Lag & 23.11 $\pm$ 9.63 & 264.59 $\pm$ 39.83 & 0.919 & 10.584 & 0.087 \\
TRPO-Lag & 6.62 $\pm$ 1.99 & 107.50 $\pm$ 37.57 & 0.263 & 4.300 & 0.062 \\
CPPO-PID & 4.20 $\pm$ 1.19 & 77.66 $\pm$ 28.87 & 0.167 & 3.106 & 0.054 \\
RCPO & 6.62 $\pm$ 1.99 & 107.50 $\pm$ 37.57 & 0.263 & 4.300 & 0.062 \\
CPO & 4.23 $\pm$ 1.87 & 61.26 $\pm$ 29.04 & 0.168 & 2.450 & 0.069 \\
PCPO & 3.13 $\pm$ 1.63 & 51.65 $\pm$ 24.37 & 0.124 & 2.066 & 0.061 \\
CUP & 5.16 $\pm$ 1.02 & 96.07 $\pm$ 41.97 & 0.205 & 3.843 & 0.054 \\
FOCOPS & 7.33 $\pm$ 1.24 & 94.09 $\pm$ 47.18 & 0.291 & 3.764 & 0.078 \\
POMDP & 9.53 $\pm$ 2.46 & 237.23 $\pm$ 31.32 & 0.379 & 9.489 & 0.040 \\
\textbf{RiskGated} & 10.86 $\pm$ 6.75 & 106.47 $\pm$ 51.19 & 0.432 & 4.259 & \textbf{0.102} \\
\bottomrule
\end{tabular}%
}
\end{minipage}
\hfill
\begin{minipage}[t]{0.40\textwidth}
\centering
\vspace{0pt}

\includegraphics[width=0.85\linewidth]{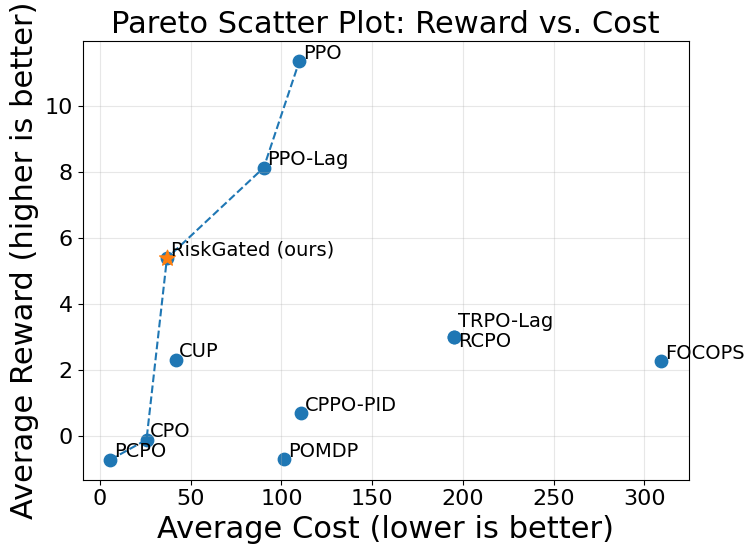}

\vspace{1mm}
{\scriptsize (a) SafetyPointGoal1}

\vspace{3mm}

\includegraphics[width=0.85\linewidth]{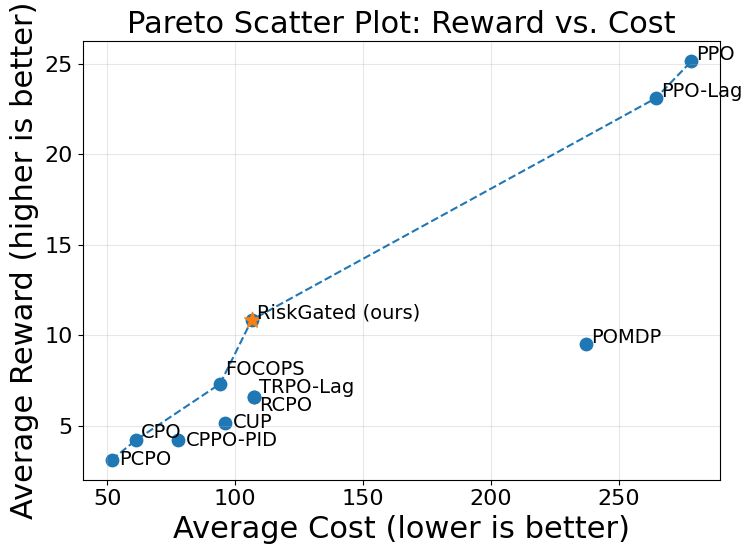}

\vspace{1mm}
{\scriptsize (b) SafetyPointCircle1}

\captionof{figure}{Pareto-style reward-cost comparison on the Safety-Gym benchmarks.}
\label{fig:safety_pareto}
\end{minipage}

\end{figure*}

\paragraph{Safety-Gym Navigation.} We next evaluate whether the risk-gated control principle transfers from glucose regulation to safety-constrained navigation. Table~\ref{tab:safety_benchmark} shows that RiskGated improves reward--cost efficiency on both Safety-Gym tasks. On SafetyPointGoal1, it reduces average cost from 109.87 to 36.93 relative to PPO while retaining nontrivial reward, yielding the highest reward-per-cost ratio among all methods. On SafetyPointCircle1, it similarly lowers cost relative to both PPO (106.47 vs.\ 278.48) and POMDP (106.47 vs.\ 237.23), and again achieves the best reward-per-cost ratio.

These results indicate that the proposed method does not maximize raw reward at all costs. Instead, it occupies a more balanced region between aggressive high-cost policies and overly conservative low-reward policies. Figure~\ref{fig:safety_pareto} visualizes this tradeoff: RiskGated lies closer to the desirable high-reward, low-cost region than unconstrained PPO-style baselines, while avoiding the reward collapse observed in several constrained safe-RL methods.

\subsection{Risk Prediction Analysis}\label{sec:risk_predict}
We next analyze whether the learned risk predictor provides a reliable decision-time signal. We compare online predicted risk with a matched post-hoc realized-pulse risk, computed using the same pulse-risk functional as the controller but replacing the learned predicted pulse with the realized next-step glucose response. Thus, this analysis tests alignment with the controller's realized risk target rather than an independent clinical ground-truth label.

\begin{figure}[htbp]
    \centering
    \includegraphics[width=0.95\linewidth]{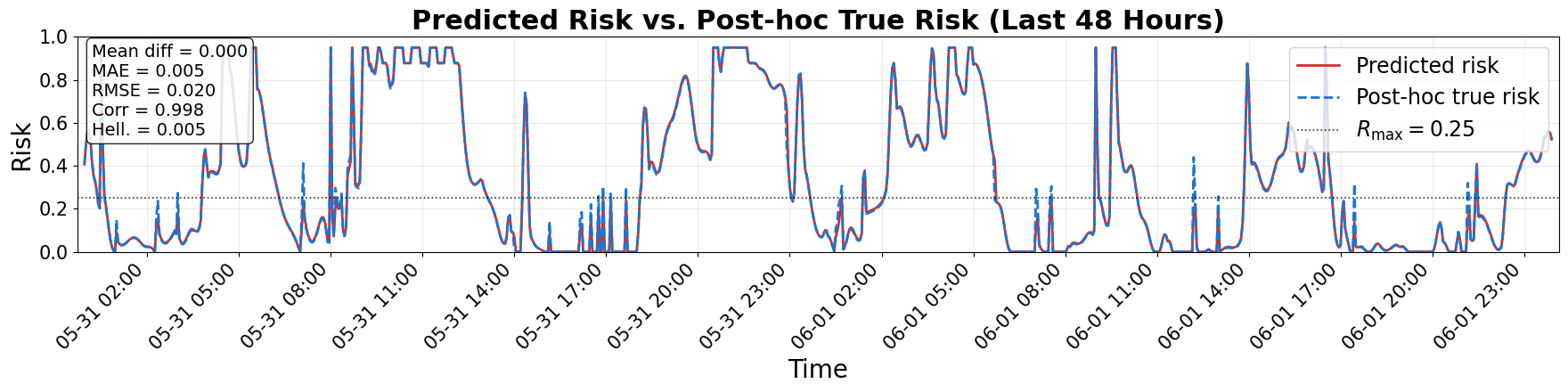}
    \caption{
    Predicted risk and matched post-hoc realized-pulse risk over the final 48 hours of evaluation.
    }
    \label{fig:risk}
\end{figure}

Figure~\ref{fig:risk} shows close alignment between predicted and post-hoc risk across both low- and high-risk periods, including near the gating threshold $R_{\max}=0.25$. Quantitatively, the predictor has mean signed error $0.0001$, MAE $0.0055$, RMSE $0.0197$, correlation $0.9982$, and Hellinger distance $0.0054$. The predicted risk exceeds $R_{\max}$ in $47.60\%$ of steps, compared with $49.27\%$ for the post-hoc risk, yielding $98.12\%$ threshold agreement. These results indicate that the predictor preserves the safe-action classification used by the risk-gated policy.

\subsection{Ablation and Sensitivity Analysis}\label{sec:sensitivity}

\paragraph{Ablation Study.}
We ablate three key design parameters: the proxy history window length $W$, the risk-admissibility threshold $R_{\max}$, and the ensemble size $M$. These parameters control the amount of recent history used for risk prediction, the size of the admissible action set, and the number of critics used for optimistic--conservative value estimation, respectively. Each configuration is evaluated over three random seeds, with mean $\pm$ standard deviation reported.

\begin{figure}[ht]
    \centering

    \begin{subfigure}[t]{0.32\textwidth}
        \centering
        \includegraphics[width=\linewidth]{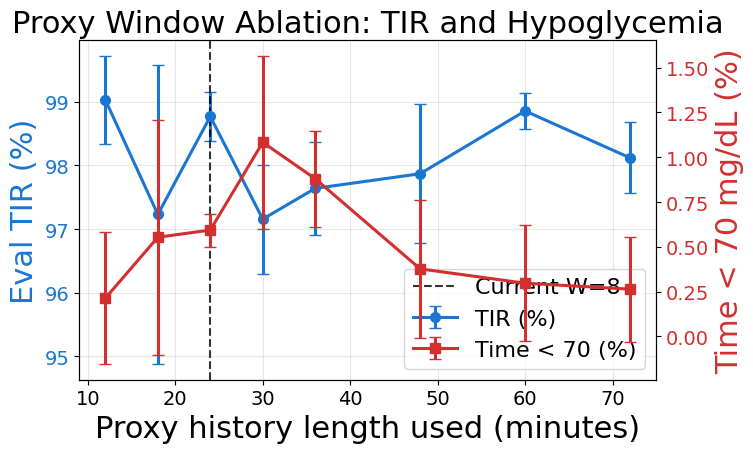}
        \caption{Proxy window $W$.}
        \label{fig:window_stability}
    \end{subfigure}
    \hfill
    \begin{subfigure}[t]{0.32\textwidth}
        \centering
        \includegraphics[width=\linewidth]{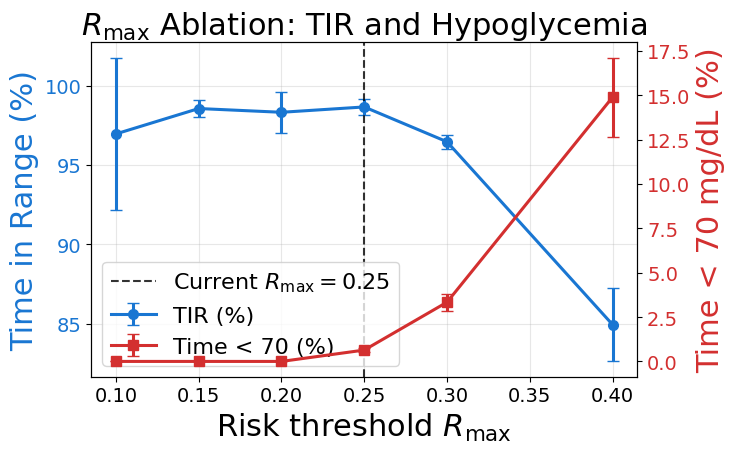}
        \caption{Risk threshold $R_{\max}$.}
        \label{fig:rmax_ablation}
    \end{subfigure}
    \hfill
    \begin{subfigure}[t]{0.32\textwidth}
        \centering
        \includegraphics[width=\linewidth]{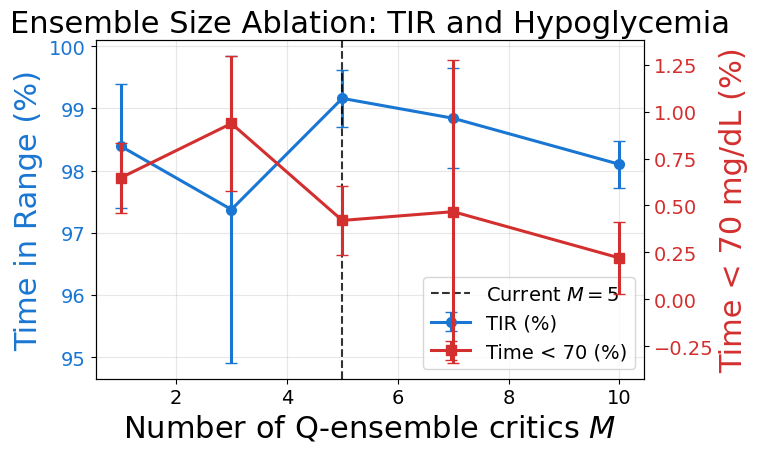}
        \caption{Ensemble size $M$.}
        \label{fig:m_ablation}
    \end{subfigure}

    \caption{
    Ablation studies for the risk-gated controller.
    (a) Sensitivity to proxy history window length $W$.
    (b) Sensitivity to risk threshold $R_{\max}$.
    (c) Sensitivity to Q-ensemble size $M$.
    Each point reports mean $\pm$ standard deviation over three random seeds.
    }
    \label{fig:window_rmax_m_ablation}
\end{figure}

Figure~\ref{fig:window_rmax_m_ablation} shows that the method is stable across nearby choices of $W$ and $M$. For the proxy-window ablation, the x-axis is reported in minutes rather than raw steps; since decisions occur every 3 minutes, the default $W=8$ corresponds to a 24-minute history window. The default values $W=8$ and $M=5$ lie in stable performance regions, indicating that the controller does not rely on finely tuned history length or ensemble size. In contrast, $R_{\max}$ has the expected safety--performance effect: smaller values impose a more conservative filter, whereas larger values admit riskier actions and can substantially increase time below 70 mg/dL. The default threshold $R_{\max}=0.25$ provides a favorable balance between TIR and hypoglycemia. Panel \ref{fig:m_ablation} shows that increasing the Q-ensemble size beyond a small number of critics does not substantially change performance: TIR remains high and hypoglycemia remains low across the tested values, with the default $M=5$ lying in a stable region. Overall, the default values $W=8$, $R_{\max}=0.25$, and $M=5$ provide a favorable balance between TIR and hypoglycemia without requiring fine tuning.

\begin{figure}[htbp]
    \centering
    \includegraphics[width=0.9\linewidth]{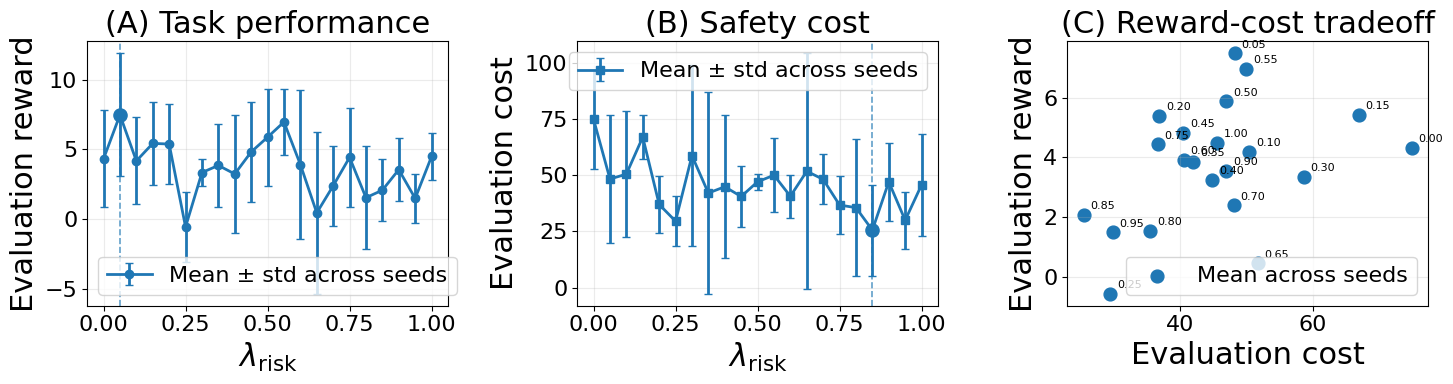}
    \caption{Sensitivity analysis of the predicted-risk penalty $\lambda_{\mathrm{risk}}$}
    \label{fig:lambda_sensitivity}
\end{figure}

\paragraph{Sensitivity to the Predicted-Risk Penalty.}
We vary $\lambda_{\mathrm{risk}}$ from 0 to 1.0 while keeping other hyperparameters fixed. Figure~\ref{fig:lambda_sensitivity} shows that $\lambda_{\mathrm{risk}}$ affects both task reward and safety cost, but not monotonically, because it only controls the soft reward-shaping penalty while the hard safety filter is still determined by the learned risk estimate and $R_{\max}$. When $\lambda_{\mathrm{risk}}=0$, the agent relies mainly on realized cost penalties and the risk-gated action filter, leading to moderate reward but relatively high cost. Intermediate values, especially around $\lambda_{\mathrm{risk}}=0.05$--$0.20$, achieve strong reward--cost tradeoffs, while larger values such as $\lambda_{\mathrm{risk}}=0.85$ place more emphasis on cost reduction. Overall, $\lambda_{\mathrm{risk}}$ acts as a conservatism knob: smaller values favor task performance, whereas larger values bias learning toward safer behavior.

\subsection{Experimental Discussion}

Across all four settings, the proposed method exhibits a consistent pattern rather than a single type of advantage. In glucose regulation, the main gains are improved overall glycemic quality together with a large computational advantage over POMDP planning: the method attains the highest TIR in both adult and adolescent cohorts while remaining competitive on hypoglycemia and hyperglycemia burden, and does so at substantially lower runtime. In Safety-Gym, the main benefit is a stronger reward-safety tradeoff: relative to PPO, the proposed method sacrifices some raw reward in exchange for much lower safety cost, while relative to POMDP and several conservative safe-RL baselines, it avoids the combination of low reward and poor efficiency. Although the exact comparison sets differ across domains, the qualitative conclusion is similar in both cases: predicted near-term hazard serves as a useful action-selection signal under partial observability, enabling the controller to adjust conservatism to the candidate action and achieve a strong balance among task performance, safety, and computational efficiency across distinct safety-critical control problems.

%% file: references.bib
@article{lauri2022pomdp,
  title={Partially Observable Markov Decision Processes in Robotics: A Survey},
  author={Lauri, Mikko and Hsu, David and Pajarinen, Joni},
  journal={IEEE Transactions on Robotics},
  year={2023},
  volume={39},
  number={1},
  pages={21--40},
  doi={10.1109/TRO.2022.3200138}
}

@inproceedings{carr2023shielding,
  title={Safe Reinforcement Learning via Shielding under Partial Observability},
  author={Carr, Steven and Jansen, Nils and Topcu, Ufuk},
  booktitle={Proceedings of the AAAI Conference on Artificial Intelligence},
  volume={37},
  number={12},
  pages={14748--14756},
  year={2023}
}

@inproceedings{pomdp,
  author    = {Cassandra, Anthony R.},
  title     = {A Survey of POMDP Applications},
  booktitle = {Working Notes of the AAAI 1998 Fall Symposium on Planning with Partially Observable Markov Decision Processes},
  volume    = {1724},
  year      = {1998},
}

@article{roy2005belief,
  author  = {Roy, Nicholas and Gordon, Geoffrey and Thrun, Sebastian},
  title   = {Finding Approximate POMDP Solutions Through Belief Compression},
  journal = {Journal of Artificial Intelligence Research},
  volume  = {23},
  pages   = {1--40},
  year    = {2005},
}

@inproceedings{silver2010pomcp,
  author    = {Silver, David and Veness, Joel},
  title     = {Monte-Carlo Planning in Large POMDPs},
  booktitle = {Advances in Neural Information Processing Systems},
  volume    = {23},
  year      = {2010},
}

@inproceedings{pineau2003pbvi,
  author    = {Pineau, Joelle and Gordon, Geoff and Thrun, Sebastian},
  title     = {Point-Based Value Iteration: An Anytime Algorithm for POMDPs},
  booktitle = {IJCAI},
  volume    = {3},
  year      = {2003},
}

@article{shani2013survey,
  author    = {Shani, Guy and Pineau, Joelle and Kaplow, Robert},
  title     = {A Survey of Point-Based POMDP Solvers},
  journal   = {Autonomous Agents and Multi-Agent Systems},
  volume    = {27},
  number    = {1},
  pages     = {1--51},
  year      = {2013},
  doi       = {10.1007/s10458-012-9200-2},
}

@inproceedings{kurniawati2008sarsop,
  author    = {Kurniawati, Hanna and Hsu, David and Lee, Wee Sun},
  title     = {SARSOP: Efficient Point-Based POMDP Planning by Approximating Optimally Reachable Belief Spaces},
  booktitle = {Robotics: Science and Systems},
  volume    = {2008},
  year      = {2008},
}

@inproceedings{igl2018deep,
  author    = {Igl, Maximilian and others},
  title     = {Deep Variational Reinforcement Learning for POMDPs},
  booktitle = {International Conference on Machine Learning},
  publisher = {PMLR},
  year      = {2018},
}

@article{kaelbling1998planning,
  title={Planning and Acting in Partially Observable Stochastic Domains},
  author={Kaelbling, Leslie Pack and Littman, Michael L. and Cassandra, Anthony R.},
  journal={Artificial Intelligence},
  volume={101},
  number={1--2},
  pages={99--134},
  year={1998},
  publisher={Elsevier},
  doi={10.1016/S0004-3702(98)00023-X}
}

@phdthesis{yu2006approximate,
  title={Approximate Solution Methods for Partially Observable Markov and Semi-Markov Decision Processes},
  author={Yu, Huan},
  school={Massachusetts Institute of Technology},
  year={2006}
}

@article{ray2019benchmarking,
  author  = {Ray, Alex and Achiam, Joshua and Amodei, Dario},
  title   = {Benchmarking Safe Exploration in Deep Reinforcement Learning},
  journal = {arXiv preprint arXiv:1910.01708},
  year    = {2019},
}

@inproceedings{achiam2017cpo,
  author    = {Achiam, Joshua and others},
  title     = {Constrained Policy Optimization},
  booktitle = {International Conference on Machine Learning},
  publisher = {PMLR},
  year      = {2017},
}

@article{ibrahim2024reward,
  author  = {Ibrahim, Sinan and others},
  title   = {Comprehensive Overview of Reward Engineering and Shaping in Advancing Reinforcement Learning Applications},
  journal = {IEEE Access},
  volume  = {12},
  pages   = {175473--175500},
  year    = {2024},
}

@article{konighofer2025shields,
  author  = {K{\"o}nighofer, Bettina and others},
  title   = {Shields for Safe Reinforcement Learning},
  journal = {Communications of the ACM},
  volume  = {68},
  number  = {11},
  pages   = {80--90},
  year    = {2025},
}

@inproceedings{rigter2023risk,
  author    = {Rigter, Marc and Lacerda, Bruno and Hawes, Nick},
  title     = {One Risk to Rule Them All: A Risk-Sensitive Perspective on Model-Based Offline Reinforcement Learning},
  booktitle = {Advances in Neural Information Processing Systems},
  volume    = {36},
  pages     = {77520--77545},
  year      = {2023},
}

@article{FDA_simulator,
  title={The UVA/PADOVA type 1 diabetes simulator: new features},
  author={Man, Chiara Dalla and Micheletto, Francesco and Lv, Dayu and Breton, Marc and Kovatchev, Boris and Cobelli, Claudio},
  journal={Journal of diabetes science and technology},
  volume={8},
  number={1},
  pages={26--34},
  year={2014},
  publisher={SAGE Publications Sage CA: Los Angeles, CA}
}

@misc{simglucose_code,
  title        = {{Simglucose} v0.2.1},
  author       = {Xie, Jinyu},
  year         = {2018},
  howpublished = {[Online]. Available: \url{https://github.com/jxx123/simglucose}},
  note         = {Accessed: 2026-04-18}
}

@article{safety_gym,
  title={Safety-Gymnasium: A Unified Safe Reinforcement Learning Benchmark},
  author={Ji, Jiaming and Zhang, Borong and Zhou, Jiayi and Pan, Xuehai and Huang, Weidong and Sun, Ruiyang and Geng, Yiran and Zhong, Yifan and Dai, Juntao and Yang, Yaodong},
  journal={arXiv preprint arXiv:2310.12567},
  year={2023}
}

@article{ppo-lag,
  author  = {Ray, Alex and Achiam, Joshua and Amodei, Dario},
  title   = {Benchmarking Safe Exploration in Deep Reinforcement Learning},
  journal = {arXiv preprint arXiv:1910.01708},
  volume  = {7},
  number  = {1},
  pages   = {2},
  year    = {2019},
}

@article{ppo,
  author  = {Schulman, John and Wolski, Filip and Dhariwal, Prafulla and Radford, Alec and Klimov, Oleg},
  title   = {Proximal Policy Optimization Algorithms},
  journal = {arXiv:1707.06347},
  year    = {2017},
}

@inproceedings{cpo,
  author    = {Achiam, Joshua and Held, David and Tamar, Aviv and Abbeel, Pieter},
  title     = {Constrained Policy Optimization},
  booktitle = {International Conference on Machine Learning},
  pages     = {22--31},
  publisher = {PMLR},
  year      = {2017},
}

@article{pcpo,
  author  = {Yang, Tsung-Yen and Rosca, Justinian and Narasimhan, Karthik and Ramadge, Peter J.},
  title   = {Projection-Based Constrained Policy Optimization},
  journal = {arXiv preprint arXiv:2010.03152},
  year    = {2020},
}

@inproceedings{focops,
  author    = {Zhang, Yiming and Vuong, Quan and Ross, Keith},
  title     = {First Order Constrained Optimization in Policy Space},
  booktitle = {Advances in Neural Information Processing Systems},
  volume    = {33},
  pages     = {15338--15349},
  year      = {2020},
}

@article{cup,
  author  = {Yang, Long and Ji, Jiaming and Dai, Juntao and Zhang, Yu and Li, Pengfei and Pan, Gang},
  title   = {CUP: A Conservative Update Policy Algorithm for Safe Reinforcement Learning},
  journal = {arXiv preprint arXiv:2202.07565},
  year    = {2022},
}

@inproceedings{trpo,
  author    = {Schulman, John and Levine, Sergey and Moritz, Philipp and Jordan, Michael and Abbeel, Pieter},
  title     = {Trust Region Policy Optimization},
  booktitle = {International Conference on Machine Learning},
  pages     = {1889--1897},
  publisher = {PMLR},
  year      = {2015},
}

@inproceedings{cppo-pid,
  author    = {Stooke, Adam and Achiam, Joshua and Abbeel, Pieter},
  title     = {Responsive Safety in Reinforcement Learning by PID Lagrangian Methods},
  booktitle = {International Conference on Machine Learning},
  pages     = {9133--9143},
  publisher = {PMLR},
  year      = {2020},
}

@article{rcpo,
  author  = {Tessler, Chen and Mankowitz, Daniel J. and Mannor, Shie},
  title   = {Reward Constrained Policy Optimization},
  journal = {arXiv preprint arXiv:1805.11074},
  year    = {2018},
}

@article{cameron2011riskmanagement,
  title   = {A Closed-Loop Artificial Pancreas Based on Risk Management},
  author  = {Cameron, Fraser and Bequette, B. Wayne and Wilson, Darrell M. and Buckingham, Bruce A. and Lee, Hyunjin and Niemeyer, G{\"u}nter},
  journal = {Journal of Diabetes Science and Technology},
  volume  = {5},
  number  = {2},
  pages   = {368--379},
  year    = {2011},
  doi     = {10.1177/193229681100500226}
}

@article{herrero2017adaptivebolus,
  title   = {Enhancing Automatic Closed-Loop Glucose Control in Type 1 Diabetes under Announced Meals Using an Adaptive Meal Bolus Calculator},
  author  = {Herrero, Pau and Haidar, Ahmad and Reddy, Madhuri and El Sharkawy, Mohamed and Pesl, Peter and Xenou, Marina and Toumazou, Christofer and Hanusch, Juan and Pankowska, Ewa and Herrero, Patricia and Oliver, Nick and Georgiou, Pantelis and Vehi, Josep},
  journal = {Artificial Intelligence in Medicine},
  volume  = {83},
  pages   = {1--8},
  year    = {2017},
  doi     = {10.1016/j.artmed.2017.06.010}
}

@article{zhao2025safeppo,
  title   = {A Safe-Enhanced Fully Closed-Loop Artificial Pancreas Controller Based on Deep Reinforcement Learning},
  author  = {Zhao, Yan Feng and Chaw, Jun Kit and Ang, Mei Choo and Tew, Yiqi and Shi, Xiao Yang and Liu, Lin and Cheng, Xiang},
  journal = {PLOS ONE},
  volume  = {20},
  number  = {1},
  pages   = {e0317662},
  year    = {2025},
  doi     = {10.1371/journal.pone.0317662}
}

@article{tejedor2023dqn,
  title   = {Evaluating Deep Q-Learning Algorithms for Controlling Blood Glucose in In Silico Type 1 Diabetes},
  author  = {Tejedor, Miguel and Hjerde, Sigurd Nordtveit and Myhre, Jonas Nordhaug and Godtliebsen, Fred},
  journal = {Diagnostics},
  volume  = {13},
  number  = {19},
  pages   = {3150},
  year    = {2023},
  doi     = {10.3390/diagnostics13193150}
}

@inproceedings{zhou2022design,
  title={Design and validation of an open-source closed-loop testbed for artificial pancreas systems},
  author={Zhou, Xugui and Kouzel, Maxfield and Ren, Haotian and Alemzadeh, Homa},
  booktitle={2022 IEEE/ACM Conference on Connected Health: Applications, Systems and Engineering Technologies (CHASE)},
  pages={1--12},
  year={2022},
  organization={IEEE}
}

@article{mixing_process,
  title={Rates of convergence for empirical processes of stationary mixing sequences},
  author={Yu, Bin},
  journal={The Annals of Probability},
  pages={94--116},
  year={1994},
  publisher={JSTOR}
}

@inproceedings{jin2021pessimism,
  title={Is pessimism provably efficient for offline rl?},
  author={Jin, Ying and Yang, Zhuoran and Wang, Zhaoran},
  booktitle={International conference on machine learning},
  pages={5084--5096},
  year={2021},
  organization={PMLR}
}

@article{tamar2016sequential,
  title={Sequential decision making with coherent risk},
  author={Tamar, Aviv and Chow, Yinlam and Ghavamzadeh, Mohammad and Mannor, Shie},
  journal={IEEE transactions on automatic control},
  volume={62},
  number={7},
  pages={3323--3338},
  year={2016},
  publisher={IEEE}
}

@article{gottesman2019guidelines,
  title={Guidelines for reinforcement learning in healthcare},
  author={Gottesman, Omer and Johansson, Fredrik and Komorowski, Matthieu and Faisal, Aldo and Sontag, David and Doshi-Velez, Finale and Celi, Leo Anthony},
  journal={Nature medicine},
  volume={25},
  number={1},
  pages={16--18},
  year={2019},
  publisher={Nature Publishing Group US New York}
}

@inproceedings{bellemare2017distributional,
  title={A distributional perspective on reinforcement learning},
  author={Bellemare, Marc G and Dabney, Will and Munos, R{\'e}mi},
  booktitle={International conference on machine learning},
  pages={449--458},
  year={2017},
  organization={Pmlr}
}

@article{kumar2020conservative,
  title={Conservative q-learning for offline reinforcement learning},
  author={Kumar, Aviral and Zhou, Aurick and Tucker, George and Levine, Sergey},
  journal={Advances in neural information processing systems},
  volume={33},
  pages={1179--1191},
  year={2020}
}
